\documentclass[a4paper]{cas-dc}
\usepackage[numbers]{natbib}
\usepackage{newunicodechar}
\newunicodechar{♫}{\textmusicalnote}
\usepackage{float}
\def\tsc#1{\csdef{#1}{\textsc{\lowercase{#1}}\xspace}}
\tsc{WGM}
\tsc{QE}
\usepackage{caption}
\usepackage{amsmath,amssymb}
\begin{document}
\let\WriteBookmarks\relax
\def\floatpagepagefraction{1}
\def\textpagefraction{.001}
\let\printorcid\relax 
\shorttitle{}    

\shortauthors{Hong-Yu An et al.}

\title[mode = title]{HyCE-RAG: Hypergraph Chain-of-Evidence Retrieval-Augmented Generation for Explainable Multi-hop Question Answering}

\author[1]{Hong-Yu An}
\ead{[yc48116@connect.um.edu.mo](mailto:yc48116@connect.um.edu.mo)}

\author[2]{Yun-Jian Zhang}
\ead{[zhangyunjian@kaihong.com](mailto:zhangyunjian@kaihong.com)}

\author[3]{Chen-Wei Liang}
\ead{[z5537371@ad.unsw.edu.au](mailto:z5537371@ad.unsw.edu.au)}

\author[4]{Tian-Yi Zhang}
\ead{[tianyizhang0213@zju.edu.cn](mailto:tianyizhang0213@zju.edu.cn)}

\author[2]{Jian Ding}
\ead{[dj19@tsinghua.org.cn](mailto:dj19@tsinghua.org.cn)}

\author[2]{Yi-Lun Wu}
\ead{[wuyilun@kaihong.com](mailto:wuyilun@kaihong.com)}

\author[2]{Ao-Bo Li}
\ead{[Liaobo@kaihong.com](mailto:Liaobo@kaihong.com)}

\author[2]{Wei-Cong Su}
\ead{[suweicong@kaihong.com](mailto:suweicong@kaihong.com)}

\author[2]{Saifullah}
\ead{[saifullah@kaihong.com](mailto:saifullah@kaihong.com)}

\author[2,5]{Mujiangshan Wang}
\cormark[1]
\ead{[mjs.wang@siat.ac.cn](mailto:mjs.wang@siat.ac.cn)}

\cortext[cor1]{Corresponding author.}

\address[1]{State Key Laboratory of Internet of Things for Smart City (SKL-IoTSC), University of Macau, Macau}

\address[2]{Shenzhen Kaihong Digital Industry Development Co., Ltd., Shenzhen 518000, China}

\address[3]{School of Mathematics and Statistics, Faculty of Science, University of New South Wales,\linebreak Sydney, NSW 2052, Australia}

\address[4]{College of Computer Science and Technology, Zhejiang University, 38 Zheda Road, Hangzhou 310027, China}

\address[5]{Shenzhen Institute of Advanced Technology, Chinese Academy of Sciences, Shenzhen 518055, China}

\begin{abstract}
Multi hop question answering requires models to retrieve relevant information from multiple documents and connect scattered evidence into a coherent reasoning process. Traditional RAG methods mainly rely on semantic similarity based retrieval, which often fails to capture complex structural relations among entities and evidence units. Graph based RAG alleviates this limitation by introducing graph structured knowledge, but pairwise relations are still insufficient for representing higher order associations involving multiple entities, facts, and contexts. Recent hypergraph based RAG methods further improve knowledge representation through hypergraphs, yet they mainly focus on retrieval and structural representation, while explicit evidence chain construction and post retrieval reasoning remain underexplored

To address this problem, we develop HyCE-RAG, namely Hypergraph Chain of Evidence Retrieval Augmented Generation, an evidence chain aware hypergraph reasoning framework for multi hop question answering. Built upon hypergraph structured knowledge representation, HyCE-RAG organizes entities, relations, and contextual evidence into hyperedges, and performs confidence aware heuristic search to select, connect, and rank relevant evidence. The search process jointly considers semantic relevance, entity connectivity, evidence coverage, reasoning coherence, and propagated confidence scores. By constructing explicit evidence chains before generation, HyCE-RAG provides large language models with more structured and interpretable reasoning context. Experiments on a recent GraphRAG benchmark and several widely used question answering datasets show that HyCE-RAG outperforms standard RAG and existing graph based RAG methods in answer accuracy, context relevance, and faithfulness. These results suggest that HyCE-RAG offers a promising direction for the post retrieval stage of RAG, shifting it from simple context aggregation toward structured evidence reasoning.

\end{abstract}



\maketitle
\section{Introduction}

Retrieval Augmented Generation (RAG) has emerged as a prominent paradigm for augmenting large language models (LLM) with external and continuously updated knowledge, thereby mitigating knowledge staleness and factual hallucination \cite{DBLP:journals/corr/abs-2002-08909,DBLP:journals/corr/abs-2312-10997}. Most standard RAG methods adopt a retrieval paradigm based on text chunks, where documents are segmented into passages and retrieved through dense vector similarity. Although this approach is effective for retrieving locally relevant contexts, it often fails to capture the semantic structures among entities, relations, and evidence units\cite{zhao2026survey}.

This limitation becomes particularly critical in complex question answering scenarios. First, semantic similarity does not necessarily imply factual utility \cite{asai2024self}. A retrieved passage may be topically related to the query, yet lack the evidence or reasoning premises required to derive the correct answer. Second, for compositional questions that require multiple reasoning steps, the necessary evidence is often scattered across different chunks or documents \cite{yang2018hotpotqa,trivedi2023interleaving}. Since conventional RAG treats these chunks as independent context units \cite{lewis2020retrieval,karpukhin2020dense}, it lacks an explicit mechanism to model dependencies across chunks, connect dispersed evidence, and construct coherent reasoning paths. Consequently, its effectiveness is limited in tasks requiring structured evidence integration and explainable reasoning \cite{jiang2023active}.
\begin{center}
    \includegraphics[width=0.95\columnwidth]{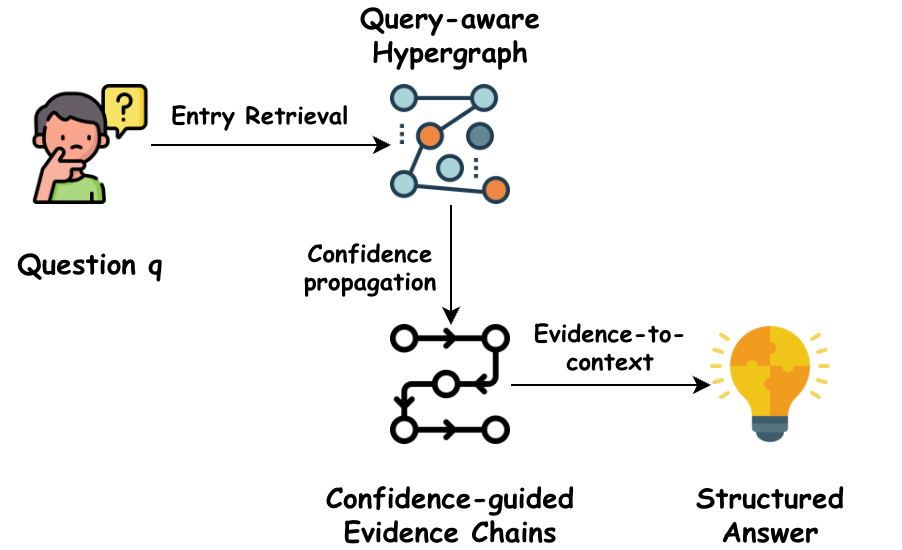}
    \captionof{figure}{Cverview of the proposed HyCE-RAG framework. It constructs a query-aware hypergraph from input question $q$, extracts confidence-guided evidence chains, and produces a structured answer.}
    \label{fig:HyCE-RAG overview}
\end{center}
To mitigate these limitations, recent studies have introduced graph-structured knowledge into RAG systems. GraphRAG \cite{edge2024local} represents entities as nodes and relations as edges, which helps organize retrieved knowledge and support knowledge-aware generation \cite{pan2024unifying}. However, conventional graph structures mainly model pairwise relations and are limited in capturing higher-order semantic associations among multiple entities, relations, and evidence units. This limitation is important for multi-hop question answering, where a single supporting paragraph may jointly connect several entities, relations, and answer-indicative facts. Encoding such evidence only as independent pairwise edges may fragment the original semantic unit and introduce spurious reasoning paths when distractor entities are present.

To address this issue, HyperGraphRAG \cite{luo2026hypergraphrag} was proposed to construct hypergraph-structured knowledge representations, where hyperedges can connect multiple entities or evidence units simultaneously. 
Prior hypergraph-based retrieval studies have shown that such high-order structures are useful for modeling complex evidence dependencies in retrieval-augmented generation. 
However, these methods largely emphasize the construction and utilization of hypergraph representations, without explicitly analyzing how retrieved evidence contributes to the final answer.
As a result, they still provide limited interpretability for post-retrieval reasoning, especially when multiple supporting and distracting evidence units are mixed together.

Motivated by these observations, we develop HyCE-RAG, namely Hypergraph Chain-of-Evidence Retrieval-Augmented Generation, an evidence-chain-aware hypergraph reasoning framework for explainable multi-hop question answering. We do not aim to re-establish the general advantage of hypergraphs over pairwise graphs, which has been investigated in prior hypergraph-based retrieval work. Instead, HyCE-RAG focuses on how confidence can be propagated and organized over a query-aware hypergraph to form faithful evidence chains for answer generation.

Unlike conventional RAG, which directly feeds retrieved text chunks into the LLM, and GraphRAG, which typically retrieves query-related nodes, edges, or subgraphs from a knowledge graph as supporting context, HyCE-RAG converts retrieved documents into a hypergraph-based evidence structure. In this structure, hyperedges connect multiple related entities and evidence units, enabling the representation of higher-order associations beyond pairwise relations. Based on this structure, HyCE-RAG performs confidence-aware heuristic search by jointly considering semantic relevance, entity connectivity, evidence coverage, reasoning coherence, and propagated confidence scores. The selected evidence is then organized into explicit evidence chains that link the question, intermediate evidence, and answer-supporting context, and these chains are provided to the LLM as structured reasoning input. Thereby, HyCE-RAG extends hypergraph-based RAG from knowledge representation to post-retrieval evidence organization and reasoning.

The main contributions of this work are summarized as follows:

\begin{itemize}
    \item We propose HyCE-RAG, Hypergraph Chain-of-Evidence Retrieval-Augmented Generation, a hypergraph-based retrieval-augmented generation framework for explainable multi-hop question answering. 
    Building on the high-order modeling capability of hypergraphs, HyCE-RAG represents entities, relations, and evidence units in a corpus-level hypergraph, enabling question-aware evidence retrieval and structured evidence organization.

    \item We introduce a confidence propagation mechanism over the query-aware hypergraph. 
    By propagating query relevance through entity--hyperedge incidence relations, HyCE-RAG identifies not only initially retrieved entry entities but also structurally connected and question-relevant entities that are important for multi-hop reasoning.

    \item We develop a confidence-guided evidence assembly and scoring strategy that organizes retrieved hyperedges, entities, and directed relations into compact evidence paths. 
    The scoring scheme jointly considers semantic relevance, entity coverage, relation reliability, and propagated confidence, allowing HyCE-RAG to select more reliable and interpretable evidence for downstream answer generation.
\end{itemize}

The remainder of this paper is organized as follows. Section 2 reviews related work. Section 3 introduces the HyCE-RAG framework and its main components. Section 4 presents the experimental results and discussion. Section 5 concludes the paper.

\section{Related Work}

\subsection{Retrieval-Augmented Generation}

Retrieval-Augmented Generation (RAG) enhances LLMs by retrieving external knowledge and conditioning generation on the retrieved contexts \cite{arslan2024survey}. 
Most existing RAG systems rely on dense retrieval, where documents are segmented into passages or chunks and ranked according to their semantic similarity to the query \cite{gupta2024comprehensive}. 
This paradigm provides LLMs with access to external and query-relevant knowledge, thereby alleviating the limitations of relying solely on parametric memory.

Despite its effectiveness, standard RAG often treats retrieved chunks as independent and flat textual units \cite{zeeshan2025rag}. 
Most RAG methods treat retrieved chunks as independent and flat textual contexts by statically splitting documents, which completely disrupts the inherent relations among contiguous textual units. As a result, retrieved contexts are often presented to the LLM as loosely organized textual fragments, making it difficult to capture entity-level dependencies, connect dispersed evidence, and form coherent reasoning paths.
This issue is especially critical in complex question answering, where deriving the correct answer often requires integrating multiple pieces of evidence distributed across different documents \cite{jiang2025retrieve}. 
These challenges have motivated structure-aware RAG methods that explicitly model relations among retrieved evidence.

\subsection{Graph-Based and Hypergraph-Based RAG}

To overcome the limitations of flat chunk retrieval, graph based RAG methods introduce structured knowledge representations into the retrieval and generation process \cite{xiang2025use}. Instead of retrieving isolated text chunks, these methods extract entities and relations from documents to construct knowledge graphs or other graph-structured indexes, and then retrieve query-related nodes, edges, subgraphs, or community summaries as generation contexts \cite{zhu2025knowledge}. By organizing knowledge through explicit graph structures, RAG based on knowledge graphs can better capture entity-level relations and provide more structured contexts for LLMs.
The use of graph structures for modeling complex connectivity, orientation, diagnosability, and fault-tolerant information transmission has a long tradition in graph theory and network science~\cite{mu2010ordered,zhao2017algorithm,wang2016diagnosability,wang2018edge,wang2011embedding}. Representative approaches in this direction include GraphRAG \cite{edge2024local}, which constructs entity-centric graphs with community summaries to support both global and local retrieval; LightRAG \cite{guo2024lightrag}, which enhances retrieval efficiency via lightweight graph indexing; and GFM-RAG \cite{luo2026gfm}, which further investigates the application of graph foundation models to retrieval-augmented generation. Beyond retrieval-specific methods, LLMs have also been leveraged to enhance knowledge graph completion~\cite{xu2024multi}, highlighting the broader potential of LLMs in improving graph-based knowledge structures.
Recent graph learning studies further show that graph neural and attention-based models can capture non-stationary, relational, and spatio-temporal dependencies in complex systems~\cite{wei2025fstgat}, which is consistent with the motivation of using structured evidence dependencies in graph-based and hypergraph-based RAG.

\begin{center}
    \includegraphics[width=0.95\columnwidth]{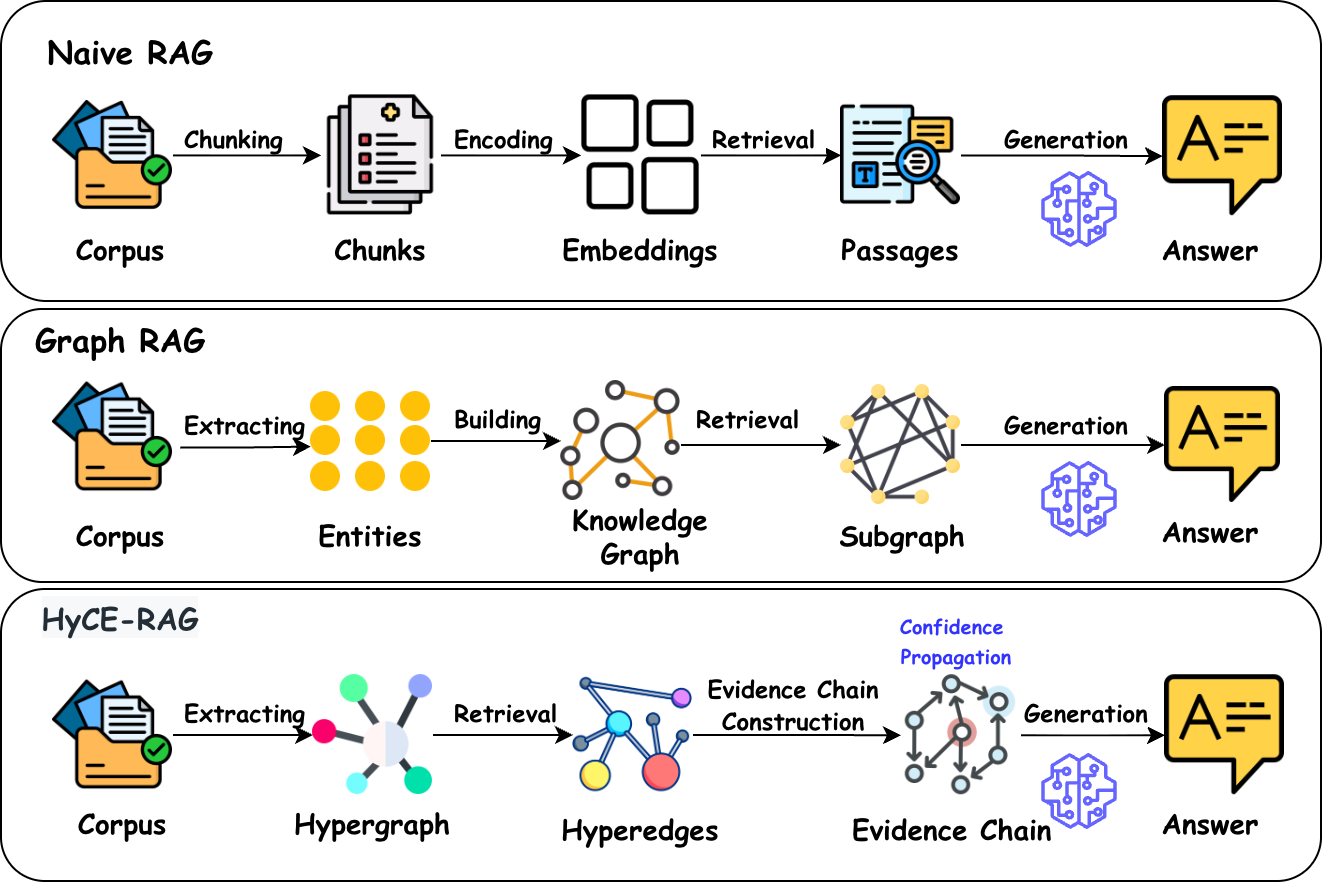}
    \captionof{figure}{Conceptual comparison among Naive RAG, GraphRAG, and HyCE-RAG. HyCE-RAG constructs confidence-guided evidence chains over hypergraph representations for structured post-retrieval reasoning.}
    \label{fig:rag_comparison}
\end{center}
However, conventional graphs mainly represent pairwise relations between entities, which may be insufficient for complex reasoning scenarios where multiple entities, events, relations, or evidence units jointly support the answer. In multi-hop question answering, a supporting paragraph often connects several entities and facts at the same time. Representing such evidence only as independent pairwise edges may fragment the original semantic unit and introduce noisy or spurious reasoning paths, especially when distractor contexts contain lexically related entities.

To address this limitation, Luo et al. \cite{luo2026hypergraphrag} introduced HyperGraph-RAG, a pioneering hypergraph-based RAG framework that uses hyperedges to connect multiple related units simultaneously. Compared with ordinary graph structures, hypergraphs provide a more expressive representation for modeling higher-order associations among entities, relations, and evidence units. Prior hypergraph-based retrieval studies suggest that high-order structures are beneficial for organizing complex evidence dependencies in retrieval-augmented generation.

Nevertheless, existing graph-based and hypergraph-based RAG methods primarily emphasize knowledge organization and retrieval, but provide limited support for explaining how the retrieved evidence is connected, weighted, and used to derive the final answer. 
This limitation is especially critical in multi-hop question answering, where models must integrate scattered evidence across multiple contexts, assess competing evidence, and form a coherent reasoning chain \cite{yang2018hotpotqa,ho2020constructing}.

\subsection{Multi-Hop Evidence Retrieval and Reasoning}

Multi-hop question answering requires models to retrieve, organize, and integrate multiple pieces of evidence before deriving the final answer \cite{mavi2024multi}. Unlike single-hop question answering, where a single passage may contain sufficient information, multi-hop questions usually involve scattered supporting facts distributed across different documents or contexts \cite{chen2019multi}. This setting poses several key challenges for retrieval-augmented systems:

\begin{enumerate}
    \item \textbf{Retrieving implicit intermediate evidence.}
    Multi-hop questions often require intermediate entities or relations that are not explicitly mentioned in the query. 
    Such evidence may have weak lexical or semantic similarity to the question, making it difficult for standard retrievers to identify all necessary supporting contexts.

    \item \textbf{Distinguishing supporting evidence from distractors.}
    Retrieved contexts may contain passages that are topically related to the query but irrelevant to the final reasoning path. 
    In some cases, they may even introduce competing or misleading evidence, requiring the system to assess evidence reliability rather than relying only on retrieval relevance.

    \item \textbf{Organizing evidence into coherent reasoning chains.}
    Even when relevant evidence is retrieved, it must be connected through entities and relations to form an interpretable reasoning path. 
    Simply concatenating loosely related context chunks provides limited support for faithful multi-hop reasoning and makes it difficult to explain how the final answer is derived.
\end{enumerate}

Existing studies have explored several strategies for multi-hop evidence retrieval and reasoning. 
Multi-hop dense retrieval methods retrieve supporting documents in multiple steps by using previously retrieved evidence to guide subsequent retrieval \cite{xiong2020answering}. 
Reasoning-path retrieval methods further aim to identify explicit evidence paths over textual or graph-structured corpora, making the retrieval process more interpretable \cite{asai2019learning}. 
More recently, iterative retrieval-and-reasoning approaches interleave evidence acquisition with intermediate reasoning, allowing LLMs to dynamically seek additional information for knowledge-intensive multi-step questions \cite{trivedi2023interleaving,yao2022react}. 
These studies collectively show that complex question answering benefits from a tighter interaction between retrieval and reasoning.

Despite this progress, most existing methods still organize evidence as sequential passages, pairwise links, or linear reasoning paths. 
Such representations are often insufficient when an answer depends on multiple evidence units that jointly support an inference. 
Moreover, once evidence is retrieved, the post-retrieval stage typically offers limited support for estimating evidence reliability, suppressing distracting contexts, and explaining how different evidence units are connected to support the final answer.

In contrast, HyCE-RAG focuses on confidence-aware post-retrieval reasoning over structured evidence. 
It leverages the high-order modeling capability of hypergraphs to organize retrieved entities, relations, and evidence units into hypergraph-based evidence chains, and propagates confidence over this structure to identify more reliable answer-supporting evidence. 
This design enables HyCE-RAG to better capture joint evidence dependencies, down-weight off-path or conflicting contexts, and provide more interpretable reasoning support for multi-hop question answering.

Related ideas also appear in long-horizon decision-making and vision--language--action reasoning, where models must maintain structured intermediate states and connect perception, language, and action evidence over extended reasoning steps~\cite{jian2026pi}.
Although the task setting is different from retrieval-augmented question answering, both lines of work emphasize the importance of structured intermediate representations for reliable multi-step reasoning.

\section{Methodology}

\subsection{Overview}

HyCE-RAG consists of two main stages: offline hypergraph system construction and online retrieval-augmented generation. 
Inspired by the hypergraph construction paradigm of HyperGraphRAG \cite{luo2026hypergraphrag}, HyCE-RAG constructs a persistent hypergraph in the offline stage.
The document corpus $\mathcal{D}$ is first divided into text chunks. 
For each chunk, an LLM-based extraction module identifies entities, hyperedges, and the incidence relations between them. 
The extracted entities and hyperedges are then deduplicated, merged, and organized into a corpus-level hypergraph $\mathcal{K}$, in which vertices denote entities or evidence units and hyperedges represent higher-order semantic associations. 
Meanwhile, an entity-level vector index is built from the names and descriptions of extracted entities, which supports semantic entry retrieval in the online stage.

In the online stage, HyCE-RAG performs question answering over the pre-built hypergraph system. 
Given a question $q$, it first identifies query-relevant entry entities by combining LLM-based query entity extraction and vector-based entity retrieval. 
Starting from these entry entities, HyCE-RAG retrieves related hyperedges and assembles a query-aware evidence hypergraph. 
It then propagates confidence scores over the hypergraph, constructs multiple candidate evidence chains, distinguishes supporting and potentially contradictory chains, and finally generates an answer using the structured evidence context.

The overall workflow can be summarized as:
\begin{equation}
\small
\begin{aligned}
\mathcal{K} 
&= \mathrm{Build}(\mathcal{D}),\\
\mathcal{V}_0 
&= \mathrm{Entry}(q,\mathcal{K}),\\
\mathcal{H}_q 
&= \mathrm{Retrieve}(\mathcal{V}_0,\mathcal{K}),\\
\mathbf{c}^{*} 
&= \mathrm{Propagate}(\mathcal{H}_q,q),\\
(\mathcal{C}_q^{+},\mathcal{C}_q^{-},\delta_q)
&= \mathrm{Reason}(\mathcal{H}_q,\mathbf{c}^{*},q),\\
y^{*} 
&= \mathrm{Generate}(q,\mathcal{C}_q^{+},\mathcal{C}_q^{-},\delta_q).
\end{aligned}
\label{eq:overall_workflow}
\end{equation}

Here, $\mathcal{K}$ denotes the offline constructed hypergraph evidence store, $\mathcal{V}_0$ is the set of entry entities, $\mathcal{H}_q$ is the query-aware evidence hypergraph, and $\mathbf{c}^{*}$ denotes the propagated confidence scores. 
$\mathcal{C}_q^{+}$ and $\mathcal{C}_q^{-}$ denote supporting and contradictory evidence chain sets, respectively, $\delta_q$ is the estimated contradiction score, and $y^{*}$ is the generated answer.
\begin{figure*}[t]
    \centering
    \includegraphics[width=0.95\textwidth]{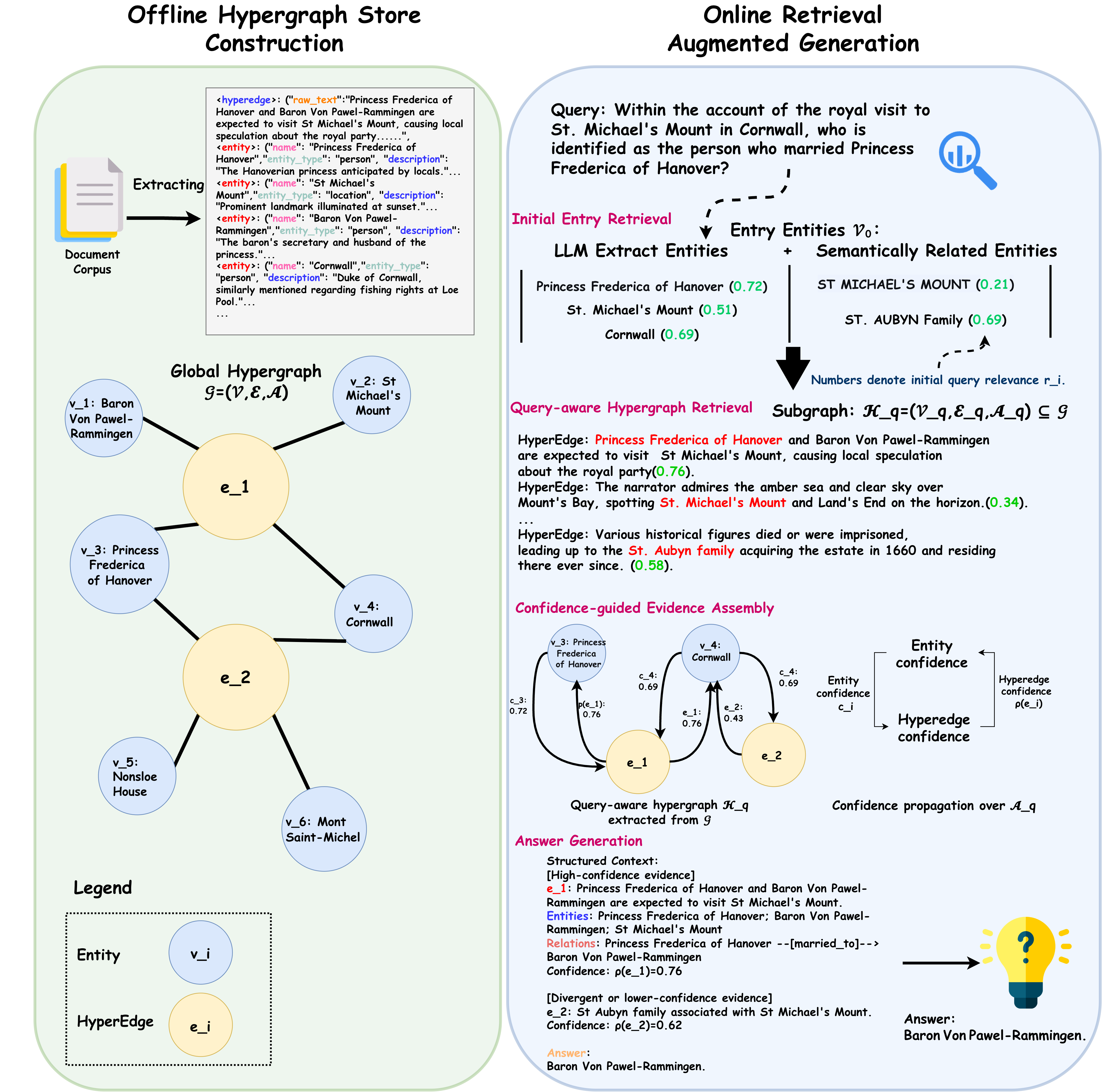}
   \caption{Overview of the proposed HyCE-RAG framework. HyCE-RAG first builds an offline hypergraph store from corpus level evidence, and then retrieves a query aware hypergraph for a given question. Confidence propagation estimates the importance of entities and hyperedges through their incidence links. Evidence units with high confidence and potential divergence are organized into a structured context for large language model answer generation.}
    \label{fig:hyce_rag_overview}
\end{figure*}
\subsection{Offline Hypergraph Store Construction}
\label{sec:offline_construction}

In this stage, HyCE-RAG converts the raw document corpus into a persistent hypergraph-based knowledge store, which is later used by the online retrieval and reasoning stage.

Given a document corpus $\mathcal{D}=\{d_i\}_{i=1}^{N}$, each document $d_i$ is segmented into a sequence of overlapping text chunks:
\begin{equation}
\mathcal{X}_i
=
\operatorname{Chunk}(d_i;L,O)
=
\{x_{i,1},x_{i,2},\ldots,x_{i,m_i}\},
\label{eq:doc_chunking}
\end{equation}
where $L$ is the maximum chunk length, $O$ is the overlap size, and $m_i$ is the number of chunks generated from document $d_i$. 
The corpus-level chunk collection is defined as:
\begin{equation}
\mathcal{X}
=
\{x_{i,k}\mid 1\leq i\leq N,\ 1\leq k\leq m_i\}.
\label{eq:corpus_chunking}
\end{equation}
Each chunk keeps its source document identifier and chunk order, enabling the extracted knowledge to be traced back to the original corpus.

For each chunk $x_{i,k}$, HyCE-RAG applies an LLM-based information extraction module to extract entities, hyperedges, and their incidence relations:
\begin{equation}
(\widehat{\mathcal{V}}_{i,k},
 \widehat{\mathcal{E}}_{i,k},
 \widehat{\mathcal{A}}_{i,k})
=
\operatorname{Extract}(x_{i,k}),
\label{eq:chunk_extraction}
\end{equation}
where $\widehat{\mathcal{V}}_{i,k}$ denotes the extracted entity records, $\widehat{\mathcal{E}}_{i,k}$ denotes the extracted hyper-relation records, and $\widehat{\mathcal{A}}_{i,k}$ denotes the incidence relations between them. 
Each extracted hyper-relation is instantiated as a hyperedge connecting the involved entities, while the source chunk is retained as provenance information for later retrieval.
Unlike a conventional graph edge that connects only two vertices, a hyperedge in HyCE-RAG can connect multiple entities within the same evidence unit, thereby representing higher-order associations among entities.

After local extraction, HyCE-RAG canonicalizes duplicated entities and equivalent hyperedges across chunks, and remaps the corresponding incidence links to global identifiers. 
The corpus-level hypergraph $\mathcal{G}$ is obtained:
\begin{equation}
\mathcal{G}
=
(\mathcal{V},\mathcal{E},\mathcal{A}),
\label{eq:global_hypergraph}
\end{equation}
where $\mathcal{V}$ is the global entity set, $\mathcal{E}$ is the global hyperedge set, and $\mathcal{A}\subseteq \mathcal{E}\times\mathcal{V}$ is the global incidence relation. 
Each hyperedge $e\in\mathcal{E}$ corresponds to an extracted hyper-relation unit and retains its source chunk information as provenance, enabling the associated textual context to be traced back to the original corpus when needed. 
The detailed record formats, canonicalization mappings, and merge operations are provided in Appendix~\ref{app:offline_details}.

To support semantic entry retrieval, HyCE-RAG further builds a vector-based entity index:
\begin{equation}
\mathcal{I}_{\mathcal{V}}
=
\{(v,\mathbf{z}_v)\mid v\in\mathcal{V},\ 
\mathbf{z}_v=f_{\mathrm{enc}}(r(v))\},
\label{eq:entity_vector_index}
\end{equation}
where $r(v)$ is the textual representation of entity $v$, and $f_{\mathrm{enc}}(\cdot)$ denotes the embedding encoder.

Finally, the offline stage produces a persistent hypergraph knowledge store:
\begin{equation}
\mathcal{K}
=
(\mathcal{G},\mathcal{I}_{\mathcal{V}}),
\label{eq:knowledge_store}
\end{equation}
where $\mathcal{G}$ supports hypergraph traversal and relation expansion, while $\mathcal{I}_{\mathcal{V}}$ supports vector-based entity retrieval.

\subsection{Online Retrieval Augmented Generation}

\subsubsection{Initial Entry Retrieval}

Given a question $q$, HyCE-RAG first identifies a set of entry entities that serve as anchors for subsequent hypergraph expansion. 
These entities are obtained from two complementary sources: explicit entity mentions extracted from the question and semantically relevant entities retrieved from the offline entity index.

Explicit entity mentions are extracted from the question using an LLM:
\begin{equation}
\mathcal{M}_q
=
\operatorname{LLMExtract}(q),
\label{eq:llm_entity_extract}
\end{equation}
where $\mathcal{M}_q$ denotes the set of entity mentions identified in $q$. 
The extracted mentions are then linked to the global entity set $\mathcal{V}$ through a deterministic linking procedure:
\begin{equation}
\mathcal{V}_{\mathrm{llm}}
=
\operatorname{Link}(\mathcal{M}_q,\mathcal{V}).
\label{eq:llm_entity_link}
\end{equation}
Here, the function $\operatorname{Link}$ first applies exact string matching against canonical entity names, and then uses alias and synonym lookup when direct matches are unavailable. 
This branch provides high-precision entry entities that are explicitly grounded in the question.

In parallel, HyCE-RAG performs semantic entity retrieval over the vector-based entity index $\mathcal{I}_{\mathcal{V}}$ constructed in the offline stage. 
For each indexed entity $v\in\mathcal{V}$, let $\mathbf{z}_v$ denote its precomputed embedding. 
The question embedding is computed as:
\begin{equation}
\mathbf{z}_q
=
f_{\mathrm{enc}}(q),
\label{eq:query_embedding}
\end{equation}
and the semantic relevance between $q$ and $v$ is measured by:
\begin{equation}
s_{\mathrm{vec}}(q,v)
=
\operatorname{sim}(\mathbf{z}_q,\mathbf{z}_v),
\label{eq:vector_entity_score}
\end{equation}
where $\operatorname{sim}(\cdot,\cdot)$ denotes the vector similarity function. 
The top-$K_v$ semantically relevant entities are selected as:
\begin{equation}
\mathcal{V}_{\mathrm{vec}}
=
\operatorname{TopK}_{v\in\mathcal{V}}^{K_v}
s_{\mathrm{vec}}(q,v).
\label{eq:vector_entity_retrieval}
\end{equation}
This branch improves recall by retrieving entities that may be relevant to the question even if they are not explicitly mentioned.

The two candidate sets are then merged and deduplicated to obtain the initial entry entity set:
\begin{equation}
\mathcal{V}_0
=
\operatorname{Dedup}
\left(
\mathcal{V}_{\mathrm{llm}}
\cup
\mathcal{V}_{\mathrm{vec}}
\right).
\label{eq:entry_entities}
\end{equation}
The resulting set $\mathcal{V}_0$ is used as the starting point for query-aware hypergraph expansion.

\subsubsection{Hyperedge Retrieval and Query-aware Hypergraph Assembly}

After obtaining the entry entity set $\mathcal{V}_0$, HyCE-RAG retrieves relevant hyperedges by expanding from these query-grounded entities over the corpus-level hypergraph. 
Let $\mathcal{E}$ denote the corpus-level hyperedge set. 
Each hyperedge $e\in\mathcal{E}$ is associated with a set of incident entities:
\begin{equation}
\mathcal{V}(e)=\{v_1,v_2,\ldots,v_m\}\subseteq \mathcal{V}, \quad m\geq 2,
\label{eq:hyperedge_def}
\end{equation}
where $m$ denotes the number of entities connected by $e$. 
The source textual segment of $e$ is retained as its provenance text, denoted as $\tau(e)$.

Instead of ranking all corpus-level hyperedges globally, HyCE-RAG first constructs a candidate hyperedge set by traversing the incidence structure from the entry entities:
\begin{equation}
\mathcal{E}_{c}
=
\{e\in\mathcal{E}\mid \mathcal{V}(e)\cap \mathcal{V}_0\neq \emptyset\},
\label{eq:candidate_hyperedges}
\end{equation}
where $\mathcal{E}_{c}$ contains hyperedges that are directly connected to at least one entry entity. 
In implementation, this expansion can be extended to multi-hop neighborhoods over the entity--hyperedge incidence graph to collect additional candidate hyperedges around the initial query entities.

For each candidate hyperedge $e\in\mathcal{E}_{c}$, HyCE-RAG estimates its query relevance from both structural and textual perspectives:
\begin{equation}
s_{\mathrm{hyp}}(q,e)
=
\beta s_{\mathrm{entry}}(e,\mathcal{V}_0)
+
(1-\beta)s_{\mathrm{rel}}(q,e),
\label{eq:hyperedge_score}
\end{equation}
where $\beta\in[0,1]$ controls the trade-off between structural entry matching and textual relevance.

The structural term $s_{\mathrm{entry}}(e,\mathcal{V}_0)$ measures how strongly the hyperedge is connected to the entry entities extracted from the question. 
It is defined as the fraction of incident entities in $e$ that overlap with $\mathcal{V}_0$:
\begin{equation}
s_{\mathrm{entry}}(e,\mathcal{V}_0)
=
\frac{|\mathcal{V}(e)\cap \mathcal{V}_0|}{|\mathcal{V}(e)|+\epsilon},
\label{eq:entry_score}
\end{equation}
where $\epsilon$ is a small constant for numerical stability. 

The textual relevance term $s_{\mathrm{rel}}(q,e)$ measures the relevance between the question $q$ and the provenance text $\tau(e)$ associated with hyperedge $e$:
\begin{equation}
s_{\mathrm{rel}}(q,e)
=
\operatorname{Rel}(q,\tau(e)),
\label{eq:textual_relevance}
\end{equation}
where $\operatorname{Rel}(\cdot)$ denotes the normalized retrieval relevance score computed between the question and the source textual segment of the hyperedge.

The query-relevant hyperedges are then selected from the candidate set:
\begin{equation}
\mathcal{E}_q
=
\operatorname{TopK}_{e\in\mathcal{E}_{c}}^{K_h}
s_{\mathrm{hyp}}(q,e),
\label{eq:hyperedge_retrieval}
\end{equation}
where $K_h$ is the number of retained hyperedges.

The query-aware vertex set is obtained by collecting all entities covered by the selected hyperedges:
\begin{equation}
\mathcal{V}_q
=
\bigcup_{e\in\mathcal{E}_q}\mathcal{V}(e).
\label{eq:query_vertices}
\end{equation}

Finally, HyCE-RAG assembles the query-aware hypergraph as:
\begin{equation}
\mathcal{H}_q
=
(\mathcal{V}_q,\mathcal{E}_q,\mathcal{A}_q),
\label{eq:query_hypergraph}
\end{equation}
where 
$\mathcal{A}_q=\mathcal{A}\cap(\mathcal{E}_q\times\mathcal{V}_q)$ denotes the incidence relations between the selected hyperedges and their participating entities. 
In $\mathcal{H}_q$, vertices represent entities, while hyperedges represent query-relevant structured units that connect multiple entities within the same textual segment. 
Compared with ordinary pairwise edges, hyperedges preserve higher-order associations among multiple entities, providing a structured basis for subsequent confidence propagation and evidence assembly.

\subsubsection{Confidence Propagation over the Hypergraph}

Given the query-aware hypergraph 
$\mathcal{H}_q=(\mathcal{V}_q,\mathcal{E}_q,\mathcal{A}_q)$, 
HyCE-RAG estimates the question-conditioned importance of entities and hyperedges through hypergraph-based confidence propagation, where $\mathcal{A}_q$ denotes the incidence relations between vertices and hyperedges. 
The intuition is that an entity should receive high confidence if it is either directly relevant to the question or structurally connected to other high-confidence entities through shared hyperedges.

Each vertex $v_i\in\mathcal{V}_q$ is first assigned an initial query relevance score:
\begin{equation}
\widehat{r}_i
=
\alpha \widetilde{s}_{\mathrm{vec}}(q,v_i)
+
(1-\alpha)\mathbf{1}(v_i\in\mathcal{V}_0),
\label{eq:query_relevance}
\end{equation}
where $\widetilde{s}_{\mathrm{vec}}(q,v_i)\in[0,1]$ is the normalized semantic similarity between the question $q$ and entity $v_i$, and $\mathbf{1}(v_i\in\mathcal{V}_0)$ indicates whether $v_i$ belongs to the initial entry entity set. 
The coefficient $\alpha\in[0,1]$ controls the trade-off between semantic relevance and explicit entry matching.

The initial scores are normalized into a restart distribution:
\begin{equation}
\mathbf{r}
=
\frac{\widehat{\mathbf{r}}}
{\|\widehat{\mathbf{r}}\|_1+\epsilon},
\qquad
\widehat{\mathbf{r}}
=
[\widehat{r}_1,\ldots,\widehat{r}_{|\mathcal{V}_q|}]^\top ,
\label{eq:init_confidence}
\end{equation}
where $\epsilon$ is a small constant for numerical stability. 
The initial confidence vector is set as $\mathbf{c}^{(0)}=\mathbf{r}$.

The incidence structure of $\mathcal{H}_q$ is encoded by 
$\mathbf{B}\in\{0,1\}^{|\mathcal{V}_q|\times|\mathcal{E}_q|}$:
\begin{equation}
\mathbf{B}_{i\ell}
=
\begin{cases}
1, & \text{if } (v_i,e_\ell)\in \mathcal{A}_q,\\
0, & \text{otherwise}.
\end{cases}
\label{eq:incidence_matrix}
\end{equation}
The vertex-degree and hyperedge-degree diagonal matrices are defined as:
\begin{equation}
\mathbf{D}_{v}(i,i)=\sum_{\ell}\mathbf{B}_{i\ell},
\qquad
\mathbf{D}_{e}(\ell,\ell)=\sum_{i}\mathbf{B}_{i\ell}.
\label{eq:degree_matrices}
\end{equation}
Isolated vertices and empty hyperedges are removed from $\mathcal{H}_q$, ensuring that all retained degrees are strictly positive.

Confidence propagation is modeled as a two-stage random walk over the hypergraph. 
Starting from a vertex, confidence is first distributed uniformly to its incident hyperedges, and then redistributed uniformly from each hyperedge to its incident vertices. 
For column-vector confidence scores, the transition matrix is:
\begin{equation}
\mathbf{S}
=
\mathbf{B}
\mathbf{D}_{e}^{-1}
\mathbf{B}^{\top}
\mathbf{D}_{v}^{-1}.
\label{eq:propagation_matrix}
\end{equation}

At propagation step $t=0,1,\ldots,K-1$, the vertex confidence vector is updated with a query-aware restart mechanism:
\begin{equation}
\mathbf{c}^{(t+1)}
=
\lambda \mathbf{r}
+
(1-\lambda)\mathbf{S}\mathbf{c}^{(t)},
\label{eq:confidence_propagation}
\end{equation}
where $\lambda\in(0,1]$ controls the restart strength. 
The restart term preserves the original query relevance signal, while the diffusion term propagates confidence through higher-order hypergraph connectivity.

Since $\mathbf{S}$ is column-stochastic and $\lambda>0$, the propagation process converges to a unique fixed point:
\begin{equation}
\mathbf{c}^{\infty}
=
\lambda
\left(
\mathbf{I}
-
(1-\lambda)\mathbf{S}
\right)^{-1}
\mathbf{r}.
\label{eq:confidence_limit}
\end{equation}
The proof is provided in Appendix~\ref{app:confidence_proof}. 
In practice, HyCE-RAG performs a finite number of propagation steps and uses 
$\mathbf{c}^{*}=\mathbf{c}^{(K)}$ as an efficient approximation to $\mathbf{c}^{\infty}$.

Finally, the confidence score of each hyperedge is computed by aggregating the final confidence scores of its incident vertices:
\begin{equation}
\rho(e_\ell)
=
\frac{1}{\mathbf{D}_{e}(\ell,\ell)}
\sum_{i=1}^{|\mathcal{V}_q|}
\mathbf{B}_{i\ell}\mathbf{c}^{*}_{i}.
\label{eq:hyperedge_confidence}
\end{equation}
The resulting vertex scores quantify question-aware entity importance, while the hyperedge scores estimate the relevance and reliability of selected hyperedges for subsequent evidence-chain construction.

\subsubsection{Confidence-guided Evidence Assembly and Multi-path Fusion}

After confidence propagation, the high-confidence region of the query-aware hypergraph is converted into compact evidence-level reasoning paths.
Each selected hyperedge corresponds to an extracted evidence unit, and its incident entities indicate the concepts involved in that evidence.
Rather than exhaustively enumerating all symbolic paths, the method constructs hyperedge-centered paths that preserve the most informative higher-order evidence associations.

For each hyperedge $e\in\mathcal{E}_q$, HyCE-RAG keeps two auxiliary extraction signals from the offline construction stage.
The first is an extraction confidence score $\gamma(e)\in[0,1]$, which reflects the reliability of the extracted evidence unit.
The second is a set of extracted directed relations $\mathcal{R}(e)$, where each relation $r\in\mathcal{R}(e)$ is associated with a confidence score $\mathrm{conf}(r)\in[0,1]$.
These signals are used only for evidence scoring and context organization, while the hypergraph structure is still defined by entities, hyperedges, and their incidence relations.

Let $\mathcal{F}_q\subseteq\mathcal{V}_q$ denote the retained high-confidence entities:
\begin{equation}
\mathcal{F}_q
=
\{v_i\in\mathcal{V}_q\mid \mathbf{c}^{*}_{i}\geq \theta_v\},
\label{eq:retained_entities}
\end{equation}
where $\theta_v\in[0,1]$ is a confidence threshold.
For each hyperedge $e\in\mathcal{E}_q$, the retained entities covered by $e$ are:
\begin{equation}
\mathcal{F}(e)
=
\{v\in\mathcal{F}_q\mid (v,e)\in\mathcal{A}_q\}.
\label{eq:covered_entities}
\end{equation}
A hyperedge is considered a candidate evidence unit if $\mathcal{F}(e)\neq\emptyset$.

Each candidate evidence unit is assigned a confidence-guided score:
\begin{equation}
\begin{aligned}
S(e)
={}&
w_{\mathrm{cov}}\mathrm{Cov}(e)
+
w_{\mathrm{prop}}\rho(e)
+
w_{\mathrm{ext}}\gamma(e)
\\
&+
w_{\mathrm{entry}}I_{\mathrm{entry}}(e)
+
w_{\mathrm{rel}}\eta_{\mathrm{rel}}(e).
\end{aligned}
\label{eq:evidence_score}
\end{equation}
Here, $\mathrm{Cov}(e)$ is the normalized coverage of retained entities, $\rho(e)$ is the propagated hyperedge confidence from Eq.~\eqref{eq:hyperedge_confidence}, and $\gamma(e)$ is the extraction confidence score.
The indicator $I_{\mathrm{entry}}(e)$ rewards evidence units that contain initial entry entities, while $\eta_{\mathrm{rel}}(e)$ measures the reliability of extracted directed relations associated with $e$.
All components are normalized to $[0,1]$, and the weights are tunable hyperparameters controlling their relative contributions.

The extraction-confidence term $w_{\mathrm{ext}}\gamma(e)$ serves as a noise-aware safeguard rather than a purely auxiliary score.
Since the hypergraph is constructed from automatically extracted entities, relations, and evidence units, extraction errors may otherwise be amplified during confidence propagation and evidence assembly.
By explicitly incorporating $\gamma(e)$ into the evidence score, HyCE-RAG down-weights evidence units whose textual extraction is uncertain, even if they cover several retained entities or receive high propagated confidence from neighboring structures.
This design reduces the risk that noisy or spurious evidence becomes a dominant reasoning path.
Meanwhile, the propagation and coverage terms still allow moderately confident evidence to be retained when it is structurally important, which balances robustness against extraction noise with the need to recover indirect multi-hop evidence.

The coverage term is defined as:
\begin{equation}
\mathrm{Cov}(e)
=
\frac{|\mathcal{F}(e)|}
{\max_{e'\in\mathcal{E}_q}|\mathcal{F}(e')|+\epsilon},
\label{eq:coverage_score}
\end{equation}
where $\epsilon$ avoids division by zero.
If no retained entity is covered by any hyperedge, the candidate evidence set is empty and the system falls back to the initially retrieved hyperedges ranked by their retrieval scores.

The entry indicator is:
\begin{equation}
I_{\mathrm{entry}}(e)
=
\mathbf{1}\left(\mathcal{V}(e)\cap\mathcal{V}_0\neq\emptyset\right),
\label{eq:entry_indicator}
\end{equation}
where $\mathcal{V}(e)=\{v\in\mathcal{V}_q\mid (v,e)\in\mathcal{A}_q\}$ denotes the set of entities incident to hyperedge $e$.

The relation-level signal is computed from the directed relations associated with $e$:
\begin{equation}
\eta_{\mathrm{rel}}(e)
=
\begin{cases}
0, & |\mathcal{R}(e)|=0,\\[4pt]
\displaystyle
\frac{1}{|\mathcal{R}(e)|}
\sum_{r\in\mathcal{R}(e)}
\mathrm{conf}(r),
& |\mathcal{R}(e)|>0,
\end{cases}
\label{eq:relation_signal}
\end{equation}
where $\mathrm{conf}(r)$ is the extraction confidence score of directed relation $r$.

The top-ranked evidence units are selected as:
\begin{equation}
\mathcal{E}_q^{*}
=
\operatorname{Top}_{L}
\left(
\{e\in\mathcal{E}_q\mid \mathcal{F}(e)\neq\emptyset\},
S(e)
\right),
\label{eq:selected_evidence_units}
\end{equation}
where $L$ is the maximum number of evidence units retained for answer generation.

For each selected hyperedge $e\in\mathcal{E}_q^{*}$, an evidence-level path is constructed as:
\begin{equation}
p_e
=
\left(
\mathcal{F}(e), e, \mathcal{R}(e), S(e), \gamma(e)
\right).
\label{eq:evidence_path}
\end{equation}
This representation records the retained entities, the corresponding hyperedge, extracted directed relations, the overall evidence score, and the extraction confidence.
Although $S(e)$ already incorporates $\gamma(e)$, the extraction confidence is kept explicitly so that the answer generator can distinguish structural relevance from extraction reliability.
For a path $p_e$, its score is defined as:
\begin{equation}
S(p_e)=S(e).
\label{eq:path_score}
\end{equation}

Since different evidence units may describe similar facts or share overlapping entities, path-level fusion is applied to reduce redundancy.
For two evidence paths $p_e$ and $p_{e'}$, their entity overlap is measured by the Jaccard coefficient:
\begin{equation}
J(p_e,p_{e'})
=
\frac{|\mathcal{F}(e)\cap\mathcal{F}(e')|}
{|\mathcal{F}(e)\cup\mathcal{F}(e')|}.
\label{eq:path_jaccard}
\end{equation}
Paths with $J(p_e,p_{e'})\geq\tau$ are regarded as highly overlapping, where $\tau$ is an overlap threshold.

HyCE-RAG uses a greedy clustering procedure for path fusion.
All paths are first sorted in descending order of $S(p)$.
The highest-scoring unassigned path is used as a seed, and all remaining unassigned paths whose Jaccard overlap with the seed is at least $\tau$ are assigned to the same group.
The grouped paths are then marked as assigned, and the procedure continues with the next highest-scoring unassigned path.
Within each group, the highest-scoring path is retained as the representative.
This procedure ensures that each path belongs to exactly one group.
The resulting compact set of representative evidence paths is denoted as $\mathcal{P}_q$.

The representative paths are further separated according to their relative confidence:
\begin{equation}
\mathcal{P}_q^{+}
=
\{p\in\mathcal{P}_q\mid S(p)\geq \theta_s S_{\max}\},
\qquad
\mathcal{P}_q^{-}
=
\mathcal{P}_q\setminus\mathcal{P}_q^{+},
\label{eq:path_partition}
\end{equation}
where $S_{\max}=\max_{p\in\mathcal{P}_q}S(p)$ and $\theta_s\in[0,1]$ is a relative confidence threshold.
The set $\mathcal{P}_q^{+}$ contains the main supporting evidence paths, while $\mathcal{P}_q^{-}$ contains lower-confidence or potentially divergent paths.

HyCE-RAG does not aggressively discard conflicting or low-confidence evidence during retrieval.
Instead, it adopts a conservative evidence exposure strategy: high-confidence supporting paths and lower-confidence paths are preserved separately in the structured context.
This allows the LLM to rely primarily on the strongest evidence while remaining aware of weaker, uncertain, or potentially inconsistent evidence.

\subsubsection{Structured Context Construction and Answer Generation}

After evidence scoring and path-level fusion, the selected evidence is linearized into a structured context for answer generation.
Unlike direct passage concatenation, the constructed context preserves the organization of supporting paths, lower-confidence paths, extracted directed relations, evidence scores, and extraction reliability signals.

Using the path sets $\mathcal{P}_q^{+}$ and $\mathcal{P}_q^{-}$ obtained above, the structured evidence context is constructed as:
\begin{equation}
\mathcal{S}_q
=
\mathrm{Ctx}
\left(
q,
\mathcal{P}_q^{+},
\mathcal{P}_q^{-}
\right),
\label{eq:structured_context}
\end{equation}
where $\mathcal{S}_q$ is the final structured context provided to the answer generator.

The context is organized into several fields.
First, the supporting paths in $\mathcal{P}_q^{+}$ are ranked by their evidence scores and presented as the primary reasoning basis.
Each path includes its involved entities, textual evidence segment, extracted directed relations, overall evidence score, and extraction confidence.
Second, the lower-confidence paths in $\mathcal{P}_q^{-}$ are included when available, exposing weaker, uncertain, or potentially inconsistent evidence to the LLM.
This separation encourages the generator to prioritize high-confidence evidence while remaining aware of alternative or conflicting information.

The final answer is generated by conditioning the LLM on both the question and the structured evidence context:
\begin{equation}
y^{*}
=
\mathrm{LLM}
\left(
q,\mathcal{S}_q
\right).
\label{eq:answer_generation}
\end{equation}

Through this structured generation process, HyCE-RAG provides the LLM with an interpretable evidence organization rather than an unstructured collection of retrieved passages.
By integrating query-aware retrieval, confidence propagation, relation-aware evidence assembly, conservative conflict exposure, and structured context construction, HyCE-RAG supports more reliable and explainable answer generation.

\section{Experiments}

We conduct experiments to evaluate HyCE-RAG from four perspectives: answer generation performance, retrieval quality, evidence organization, and component effectiveness. Specifically, we aim to answer the following research questions:

\textbf{RQ1:} Does HyCE-RAG improve answer generation performance on multi hop question answering tasks?

\textbf{RQ2:} Does confidence propagation over the hypergraph improve the quality of retrieved evidence?

\textbf{RQ3:} How do confidence propagation and hyperedge-based evidence assembly contribute to the final performance?

\textbf{RQ4:} Can HyCE-RAG provide interpretable evidence structures for complex multi-hop question answering?

\subsection{Experimental Setup}

\subsubsection{Datasets}

To evaluate HyCE-RAG across both general and domain-specific reasoning scenarios, we conduct experiments on three established multi-hop question answering benchmarks, namely HotpotQA~\cite{yang2018hotpotqa}, 2WikiMultihopQA~\cite{ho2020constructing}, and MuSiQue~\cite{trivedi2022musique}, as well as two domain-specific subsets of the GraphRAG Benchmark~\cite{xiao2025graphrag}.
Below we summarize the key characteristics of each evaluation setting.

\noindent\textbf{HotpotQA.}
HotpotQA is a multi hop question answering dataset that requires reasoning over multiple Wikipedia passages. It contains approximately 90,000 question answer pairs under the distractor setting, in which each question is paired with both gold supporting passages and distractor passages. This setup enables evaluation of whether HyCE-RAG can identify useful evidence while suppressing irrelevant information.

\noindent\textbf{2WikiMultihopQA.}
2WikiMultihopQA contains approximately 193,000 questions, each requiring reasoning over two to four Wikipedia articles. The dataset emphasizes compositional reasoning, including comparison, inference, and bridge style reasoning across multiple entities.

\noindent\textbf{MuSiQue.}
MuSiQue is a challenging benchmark with approximately 25,000 questions, each requiring two to four reasoning steps. Unlike datasets whose questions can often be solved through shortcut reasoning, MuSiQue is designed to require explicit decomposition and coherent multi step evidence aggregation.

\noindent\textbf{GraphRAG Benchmark.}
The GraphRAG Benchmark evaluates graph based retrieval augmented generation in structured long context settings. We use both the medical and novel subsets. The medical subset focuses on factual retrieval and complex reasoning over clinical resources, including treatment guidelines and diagnostic knowledge. The novel subset requires reasoning over characters, events, temporal dependencies, and cross document narrative relations. Together, these two subsets evaluate HyCE-RAG under both knowledge intensive and narrative reasoning scenarios.

\subsubsection{Baselines}

We compare HyCE-RAG with representative baselines that cover closed-book generation, standard retrieval-augmented generation, and graph-based retrieval-augmented generation.

\noindent\textbf{No Context.}
The LLM answers each question using only its parametric knowledge, without access to any retrieved or provided external evidence. 
This setting follows the closed-book QA paradigm~\cite{roberts2020much} and serves as a lower bound for measuring how much external evidence is needed for each dataset.

\noindent\textbf{Vector-based RAG.}
This baseline retrieves passages according to dense semantic similarity between the question and candidate passages, following the standard dense retrieval and retrieval-augmented generation paradigm~\cite{karpukhin2020dense,lewis2020retrieval}. 
The top-ranked passages are concatenated and provided to the LLM for answer generation. 
It represents the standard retrieve-then-generate pipeline without explicit graph or hypergraph structure.

\noindent\textbf{GraphRAG.}
GraphRAG constructs a graph-based representation over the corpus and retrieves evidence through graph-structured search or neighborhood expansion~\cite{edge2024local}. 
This baseline evaluates whether pairwise graph relations can improve evidence retrieval compared with purely vector-based retrieval.

\noindent\textbf{LightRAG.}
LightRAG combines efficient indexing with graph-aware retrieval~\cite{guo2024lightrag}. 
It uses both semantic retrieval signals and structured associations to retrieve relevant evidence for generation, providing a strong graph-enhanced RAG comparison.

For a fair comparison, all methods use the same answer generator, prompt template, and generation parameters unless otherwise specified.

\subsubsection{Evaluation Metrics}

Following the evaluation protocols used in recent benchmarks such as GraphRAG Benchmark~\cite{xiao2025graphrag} and LinearRAG~\cite{zhuang2025linearrag}, we evaluate each method comprehensively. 
Traditional QA evaluation metrics, such as Exact Match (EM) or F1-score, primarily focus on literal string matching or the extraction of specific entities. However, in complex multi-hop reasoning scenarios, datasets often lack fine-grained ground-truth annotations for intermediate reasoning paths, making these lexical metrics inadequate and prone to false negatives. 
To more accurately assess the semantic correctness, logical coherence, and factual grounding of the models, we adopt the widely recognized LLM-as-a-judge paradigm. 
Specifically, we evaluate each method from three complementary aspects: whether the final answer is accurate, whether the retrieved context is relevant to the question, and whether the answer is supported by the retrieved context. 
These aspects are measured by answer accuracy, context relevance, and faithfulness, respectively.

\noindent\textbf{Answer Accuracy.}
Answer accuracy ($\mathrm {Acc}$) evaluates whether the generated answer is consistent with the reference answer in terms of both factual content and semantic meaning. 
Given a generated answer $a$ and a reference answer $\hat{a}$, an LLM evaluator decomposes them into atomic factual statements and compares the resulting statement sets. 
It identifies true positives ($TP$), false positives ($FP$), and false negatives ($FN$), where $TP$ denotes correctly covered reference facts, $FP$ denotes incorrect facts introduced by the generated answer, and $FN$ denotes reference facts missed by the generated answer. 
The factual matching score is computed as:
\begin{equation}
F_{\mathrm{fact}} =
\frac{TP}{TP + \frac{1}{2}(FP + FN)} .
\end{equation}
To account for semantically equivalent expressions, we further compute an embedding-based semantic similarity score $S_{\mathrm{sem}}(a,\hat{a})$. 
The final answer accuracy score is computed as:
\begin{equation}
\mathrm{Acc}
=
\alpha F_{\mathrm{fact}}
+
(1-\alpha)S_{\mathrm{sem}}(a,\hat{a}),
\end{equation}
where $\alpha$ controls the relative contribution of factual matching and semantic similarity.

\noindent\textbf{Context Relevance.}
Context relevance ($\mathrm{Rel}$) measures whether the retrieved context is useful for answering the input question. 
Given a question $q$ and the retrieved context $C$, we assess relevance from two complementary perspectives. 
First, an LLM evaluator judges whether $C$ contains information needed to answer $q$ and assigns a relevance score $\mathrm{Rel}_{\mathrm{LLM}}(q,C)$. 
Second, we compute an embedding-based semantic relevance score $\mathrm{Rel}_{\mathrm{sim}}(q,C)$ between the question and the retrieved context. 
The final context relevance score is computed as:
\begin{equation}
\mathrm{Rel}
=
\lambda \mathrm{Rel}_{\mathrm{LLM}}(q,C)
+
(1-\lambda)\mathrm{Rel}_{\mathrm{sim}}(q,C),
\end{equation}
where $\lambda$ controls the relative contribution of LLM-based relevance judgment and embedding-based semantic relevance.

\noindent\textbf{Faithfulness.}
Faithfulness ($\mathrm {Faith}$) measures whether the generated answer is grounded in the retrieved context. 
Given a generated answer $a$ and the retrieved context $C$, an LLM evaluator first decomposes $a$ into atomic factual statements. 
For each statement, the LLM evaluator determines whether it is supported by $C$. 
The faithfulness score is defined as the proportion of supported statements:
\begin{equation}
\mathrm{Faith}
=
\frac{N_{\mathrm{supported}}}{N_{\mathrm{total}}},
\end{equation}
where $N_{\mathrm{total}}$ is the total number of atomic statements extracted from the generated answer, and $N_{\mathrm{supported}}$ is the number of statements supported by the retrieved context. 
A higher faithfulness score indicates that the generated answer is better grounded in the retrieved evidence and contains fewer unsupported claims.

\subsubsection{Implementation Details}

All experiments employ \texttt{gpt-oss-120b} as the backbone language model for both hypergraph construction and answer generation, encompassing entity extraction, relation extraction, evidence hyperedge construction, and final response generation~\cite{openai2025gptoss120bgptoss20bmodel}. The model is deployed using \texttt{vLLM} to enable efficient inference~\cite{kwon2023efficient}. All experiments are conducted on a server equipped with six NVIDIA RTX 4090 GPUs.

We use \texttt{sentence-transformers/all-MiniLM-L6-v2} as the embedding model throughout all embedding-based components. Specifically, it is used to build dense indexes and encode entities, queries, passages, and document chunks for similarity search and initial evidence retrieval. 
The same embedding model is also used by the vector-based RAG baseline, ensuring that performance differences are not caused by different dense encoders.

We use a fixed set of hyperparameters for HyCE-RAG across all test examples. 
The detailed hyperparameter settings, including the number of entry entities, expansion hops, propagation steps, and selected evidence units, are reported in Appendix~\ref{app:hyperparameters}. 
For fair comparison, all retrieval-based methods are evaluated under comparable evidence budgets and input length constraints.

\subsection{Results and Discussion}

Table~\ref{tab:main_results_multihop} and Table~\ref{tab:main_results_graphragbench} report the results on multi-hop QA and GraphRAG-Bench datasets, respectively.  
Answer quality is evaluated by answer correctness, while context relevance and faithfulness measure the quality of retrieved evidence when external context is available.  
For the No Context baseline, relevance and faithfulness are not applicable, as no external evidence is retrieved.

\begin{table}[t]
\centering
\small
\caption{Main results on multi-hop QA datasets. 
All values are percentages. 
$\mathrm{Acc}$, $\mathrm{Rel}$, and $\mathrm{Faith}$ denote answer correctness, context relevance, and faithfulness. 
N/A denotes metrics not applicable to methods without retrieved context.}
\label{tab:main_results_multihop}
\resizebox{\columnwidth}{!}{
\begin{tabular}{llccc}
\toprule
\textbf{Dataset} & \textbf{Method} & \textbf{Acc} & \textbf{Rel} & \textbf{Faith} \\
\midrule
\multirow{5}{*}{HotpotQA}
& No Context       & 31.11 & N/A & N/A \\
& Vector RAG       & 48.31 & 56.27 & 54.84 \\
& GraphRAG         & 42.20 & 47.09 & 51.13 \\
& LightRAG         & 60.17 & 58.62 & 54.61 \\
& HyCE-RAG         & \textbf{71.57} & \textbf{61.54} & \textbf{60.10} \\
\midrule
\multirow{5}{*}{2Wiki}
& No Context       & 32.64 & N/A & N/A \\
& Vector RAG       & 39.52 & 48.27 & 51.11 \\
& GraphRAG         & 46.60 & 45.66 & 48.19 \\
& LightRAG         & 52.82 & 57.34 & 54.29 \\
& HyCE-RAG         & \textbf{70.97} & \textbf{66.76} & \textbf{56.09} \\
\midrule
\multirow{5}{*}{MuSiQue}
& No Context       & 17.82 & N/A & N/A \\
& Vector RAG       & 19.26 & 34.62 & 37.08 \\
& GraphRAG         & 14.94 & 28.15 & 44.57 \\
& LightRAG         & 27.12 & 34.65 & 37.22 \\
& HyCE-RAG         & \textbf{56.73} & \textbf{57.16} & \textbf{79.23} \\
\bottomrule
\end{tabular}
}
\end{table}

\begin{table}[t]
\centering
\small
\caption{Main results on GraphRAG-Bench datasets. 
All values are percentages. 
$\mathrm{Acc}$, $\mathrm{Rel}$, and $\mathrm{Faith}$ denote answer correctness, context relevance, and faithfulness. 
N/A denotes metrics not applicable to methods without retrieved context.}
\label{tab:main_results_graphragbench}
\resizebox{\columnwidth}{!}{
\begin{tabular}{llccc}
\toprule
\textbf{Dataset} & \textbf{Method} & \textbf{Acc} & \textbf{Rel} & \textbf{Faith} \\
\midrule
\multirow{5}{*}{Medical}
& No Context       & 37.31 & N/A & N/A \\
& Vector RAG       & 49.12 & 53.28 & 58.62 \\
& GraphRAG         & 53.36 & 57.64 & 69.21 \\
& LightRAG         & 54.27 & 61.27 & 67.09 \\
& HyCE-RAG         & \textbf{70.58} & \textbf{64.53} & \textbf{81.71} \\
\midrule
\multirow{5}{*}{Novel}
& No Context       & 24.29 & N/A & N/A \\
& Vector RAG       & 44.37 & 55.67 & 58.27 \\
& GraphRAG         & 48.83 & 52.19 & 56.78 \\
& LightRAG         & 51.12 & 56.09 & 61.24 \\
& HyCE-RAG         & \textbf{66.49} & \textbf{62.18} & \textbf{79.83} \\
\bottomrule
\end{tabular}
}
\end{table}

Overall, HyCE-RAG consistently achieves the best performance across all datasets and all applicable metrics. 
On the multi-hop QA benchmarks, HyCE-RAG improves answer correctness by 11.4, 18.2, and 29.6 absolute points over the strongest baseline on HotpotQA, 2Wiki, and MuSiQue, respectively. 
On GraphRAG-Bench, it also outperforms all baselines, with accuracy gains of 16.3 points on Medical and 15.4 points on Novel. 
These results demonstrate that HyCE-RAG provides more effective evidence for complex question answering than both dense vector retrieval and existing graph-based retrieval methods.

Prior studies~\cite{xiang2025use} have shown that graph-based RAG can be beneficial for multi-hop question answering because graph search captures relations among documents and entities, rather than treating retrieved chunks as an independent ranked list. 
However, our results indicate that graph construction alone is not sufficient. 
GraphRAG does not consistently outperform the vector-based baseline, especially on HotpotQA and MuSiQue, suggesting that simply summarizing or aggregating graph communities may introduce noisy or weakly relevant evidence. 
This issue becomes more serious when the retrieved context contains both supporting and distracting information.

The advantage of HyCE-RAG is most evident on MuSiQue, where the task requires reasoning over multiple connected pieces of evidence while filtering out misleading contexts. 
HyCE-RAG improves context relevance from 34.65\% to 57.16\% and faithfulness from 44.57\% to 79.23\% over the strongest corresponding baselines, corresponding to relative gains of 65.0\% and 77.8\%, respectively.

This large improvement suggests that directly extracting a subgraph is not the final goal; the system must further identify the reasoning thread within the retrieved structure. 
HyCE-RAG addresses this problem through confidence propagation over entities and hyperedges, followed by relation-aware evidence scoring.
This design assigns higher scores to evidence units that are structurally connected, semantically relevant, and supported by reliable extracted relations.
A qualitative case study in Section~\ref{sec:case_study} further illustrates how such propagation helps HyCE-RAG identify a coherent evidence path for multi-hop reasoning.

Compared with LightRAG, HyCE-RAG further benefits from evidence-level confidence estimation rather than relying only on graph-guided retrieval. 
The propagated confidence scores help prioritize coherent evidence paths before generation, allowing the language model to condition on more reliable and informative context. 
As a result, HyCE-RAG produces answers with both higher correctness and stronger grounding, as reflected by the consistent improvements in faithfulness across all datasets.

In addition, we note that the construction of the hypergraph is performed in an offline stage. 
Although this stage may involve LLM-based information extraction, it is a one-time preprocessing cost for a fixed corpus. 
During online inference, HyCE-RAG mainly performs retrieval, confidence propagation, and evidence scoring over a sparse hypergraph structure, which can be efficiently implemented with sparse matrix operations. 
Therefore, the additional structural modeling cost does not substantially change the online generation pipeline.

\subsection{Ablation Study}
\label{sec:ablation}

To answer RQ3, we conduct an ablation study to examine the contribution of the core components in HyCE-RAG.
Due to the computational cost of reconstructing hypergraph indexes and rerunning generation, we perform ablations on MuSiQue, a challenging multi-hop QA dataset that requires reasoning over multiple connected pieces of evidence.
This setting is suitable for evaluating whether HyCE-RAG can organize and retrieve coherent evidence for multi-hop reasoning.
All variants are evaluated under the same evidence budget and generation settings as the full HyCE-RAG model.

We compare HyCE-RAG with the following ablated variants.

\noindent\textbf{HG-RAG.}
HG-RAG uses the same hypergraph construction as HyCE-RAG but removes both confidence propagation and hyperedge-based evidence assembly.
It ranks hypergraph evidence only by initial query relevance before generation.
This variant examines whether hypergraph construction alone is sufficient for reliable multi-hop evidence retrieval.

\noindent\textbf{w/o Confidence Propagation.}
This variant removes confidence propagation over the hypergraph.
Evidence units are selected only according to their initial query relevance scores, without propagating confidence through connected entities and hyperedges.

This reduces HyCE-RAG to a direct relevance-based selection strategy and evaluates whether hypergraph propagation helps recover indirect multi-hop evidence that is not strongly matched by the query alone.

\noindent\textbf{w/o Hyperedge Evidence Assembly.}
This variant removes hyperedge-based evidence assembly.
Instead of grouping related entities and facts into evidence hyperedges, the system treats evidence units as independent textual pieces and ranks them directly according to their relevance scores.
This setting evaluates whether higher-order evidence organization helps preserve coherent reasoning contexts for answer generation.

\begin{table}[t]
\centering
\caption{Ablation results on MuSiQue. 
All values are percentages. 
$\mathrm{Acc}$, $\mathrm{Rel}$, and $\mathrm{Faith}$ denote answer correctness, context relevance, and faithfulness, respectively.}
\label{tab:ablation}
\resizebox{\columnwidth}{!}{
\begin{tabular}{lccc}
\toprule
\textbf{Method} & \textbf{Acc} & \textbf{Rel} & \textbf{Faith} \\
\midrule
HG-RAG                         & 24.36 & 27.81 & 33.79 \\
w/o Confidence Propagation     & 37.22 & 41.67 & 43.56 \\
w/o Hyperedge Evidence Assembly & 31.57 & 36.28 & 39.12 \\
\midrule
HyCE-RAG                       & \textbf{56.73} & \textbf{57.16} & \textbf{79.23} \\
\bottomrule
\end{tabular}
}
\end{table}

Table~\ref{tab:ablation} shows that removing any core component degrades the performance of HyCE-RAG on MuSiQue.
HG-RAG performs the worst among all variants, indicating that hypergraph construction alone is insufficient for effective multi-hop evidence retrieval.
Although the hypergraph provides a structured representation of entities and relations, reliable reasoning still requires mechanisms that can identify structurally important evidence and assemble it into coherent contexts.

Without confidence propagation, the model relies only on direct query--evidence relevance.
This is less effective for questions whose supporting facts are connected through intermediate entities or relations.
The result suggests that propagating confidence over the hypergraph helps identify evidence units that are not individually prominent but are structurally important for multi-hop reasoning.

Removing hyperedge-based evidence assembly also leads to a substantial performance drop.
This indicates that treating evidence as isolated textual units may break the connections among related facts and produce less coherent retrieval contexts.
By grouping entities, relations, and supporting facts into hyperedges, HyCE-RAG provides the generator with more complete evidence structures, which contributes to both answer correctness and faithfulness.

\subsection{Case Study}
\label{sec:case_study}

To further illustrate the reasoning behavior of HyCE-RAG, we present a case study from the four-hop subset of MuSiQue.\footnote{Question ID: \texttt{4hop1\_\_39871\_314549\_131976\_90181}.}
Each MuSiQue instance contains 20 context paragraphs, among which only 4 are supporting paragraphs and the remaining 16 are distractor paragraphs.
The question is:

\begin{quote}
\small
\emph{What is the deepest part of the ocean by the state where Main Street Station is located?}
\end{quote}

The ground-truth answer is \textit{Milwaukee Deep}.
This example requires the system to recover a four-hop reasoning chain from the entity mentioned in the question to the relevant geographic region, the corresponding ocean, and finally its deepest point.
The setting is challenging because most context paragraphs are distractors, and some retrieved evidence is lexically related to the question but does not lie on the correct reasoning path.

\begin{table}[t]
\centering
\caption{Case study on a four-hop MuSiQue question. HyCE-RAG assembles evidence chains that support the answer \textit{Milwaukee Deep} while suppressing a structurally inconsistent chain. Green (red) tags denote supporting (contradicting) chains.}
\label{tab:case_study}
\small
\renewcommand{\arraystretch}{1.5}
\begin{tabular}{@{}p{0.18\columnwidth} p{0.64\columnwidth} c@{}}
\toprule
\textbf{Role} & \textbf{Assembled Evidence Chain} & \textbf{Score} \\
\midrule

\cellcolor{green!8}Answer support
& \cellcolor{green!8}\textbf{Final answer:} \textit{Milwaukee Deep} \newline
\textit{Milwaukee Deep}
\,$\xrightarrow{\text{\scriptsize is\_part\_of}}$\,
\textit{Atlantic Ocean}
\,$\xrightarrow{\text{\scriptsize associated\_with}}$\,
\textit{8,648 metres}
& \cellcolor{green!8}0.8617 \\

\cmidrule(lr){2-3}

\cellcolor{green!8}Path support
& \cellcolor{green!8}\textit{Puerto Rico Trench}
\,$\xrightarrow{\text{\scriptsize associated\_with}}$\,
\textit{Atlantic Ocean}
\,$\xrightarrow{\text{\scriptsize associated\_with}}$\,
\textit{Caribbean Sea}
& \cellcolor{green!8}0.8062 \\

\cmidrule(lr){2-3}

\cellcolor{green!8}Context support
& \cellcolor{green!8}\textit{Atlantic Ocean}
\,$\rightarrow$\,
\textit{Trujillo Alto}
\,$\rightarrow$\,
\textit{Guaynabo}
\,$\rightarrow$\,
\textit{Carolina}
\,$\rightarrow$\,
\textit{Caguas}
& \cellcolor{green!8}0.7442 \\

\midrule

\cellcolor{red!8}Disagreement
& \cellcolor{red!8}\textbf{Lexically related but off-path:} \newline
\textit{G Train}
\,$\xrightarrow{\text{\scriptsize stop\_at}}$\,
\textit{Main Street Station}
\,$\xrightarrow{\text{\scriptsize located\_in}}$\,
\textit{Williamsburg}
& \cellcolor{red!8}0.4285 \\

\bottomrule
\end{tabular}
\end{table}

This case demonstrates two key advantages of HyCE-RAG: confidence propagation and structured evidence assembly.
First, confidence propagation enables HyCE-RAG to move beyond direct lexical matching.
Instead of ranking evidence solely by its initial query similarity, HyCE-RAG propagates confidence over connected entities, hyperedges, and relations.
As shown in Table~\ref{tab:case_study}, evidence chains that are structurally consistent with the correct reasoning path receive higher propagated scores.
For instance, the chain connecting \textit{Milwaukee Deep} with the \textit{Atlantic Ocean} obtains the highest score of 0.8617, directly supporting the final answer.
The chain involving the \textit{Puerto Rico Trench} and the \textit{Atlantic Ocean} also receives a high score of 0.8062, further reinforcing the geographic reasoning path.
In contrast, the chain involving \textit{G Train} and \textit{Williamsburg} receives a substantially lower score of 0.4285.
Although this evidence is lexically related to entities in the retrieved context, it is not structurally aligned with the target ocean-depth reasoning path.

Second, structured evidence assembly allows HyCE-RAG to organize scattered pieces of evidence into coherent reasoning units.
Rather than presenting isolated passages to the generator, HyCE-RAG assembles related entities and relations into evidence chains that explicitly expose how different facts support the answer.
This organization helps the generator focus on the evidence leading to \textit{Milwaukee Deep}, while reducing the influence of off-path or conflicting evidence.
In this example, the supporting chains jointly connect the final answer, the relevant ocean, and related geographic clues, whereas the \textit{G Train}--\textit{Williamsburg} chain is down-weighted because it does not contribute to the required reasoning trajectory.

Overall, this case suggests that HyCE-RAG not only retrieves relevant evidence, but also differentiates structurally supportive evidence from misleading or off-path evidence.
By combining confidence propagation with evidence-chain assembly, HyCE-RAG provides the generator with a more reliable and interpretable context, leading to the correct answer, \textit{Milwaukee Deep}.

\section{Conclusion}
\label{sec:conclusion}

We presented HyCE-RAG, a hypergraph-based retrieval-augmented generation framework for complex question answering. 
Instead of ranking passages independently, HyCE-RAG models entities, relations, and evidence units within a unified hypergraph and applies confidence propagation to identify structurally coherent evidence chains. 
This design helps the generator focus on evidence that is both semantically relevant and structurally supportive, while reducing the influence of off-path distractors.

Experiments on multi-hop QA and GraphRAG-Bench datasets show that HyCE-RAG achieves stronger overall performance than competitive retrieval-based baselines.
Averaged across the five datasets, HyCE-RAG improves over the strongest corresponding baseline for each metric by 18.17 percentage points in answer correctness, 8.84 percentage points in context relevance, and 14.56 percentage points in faithfulness.
These results indicate that hypergraph-structured evidence organization is effective for improving both retrieval quality and answer grounding, especially in settings that require multi-hop reasoning over dispersed evidence.

Future work will explore more efficient hypergraph construction and extend HyCE-RAG to broader domain-specific and long-context reasoning scenarios.

\section*{Appendix}
\addcontentsline{toc}{section}{Appendix}

\appendix

\section{Details of Offline Hypergraph Store Construction}
\label{app:offline_details}

This appendix provides the implementation-level details of the offline hypergraph store construction procedure described in Section~\ref{sec:offline_construction}, including the record formats, canonicalization mappings, merge operations, and storage implementation.

Each hyper-relation record extracted from a chunk is instantiated as a hyperedge. 
Specifically, a hyperedge $e\in\widehat{\mathcal{E}}_{i,k}$ corresponds to a self-contained structured knowledge segment and is represented as:
\begin{equation}
e
=
(\mathrm{seg},\mathrm{time},\mathrm{src},\mathrm{doc},\mathrm{chunk}),
\label{eq:app_hyperedge_record}
\end{equation}
where $\mathrm{seg}$ denotes the extracted knowledge segment, $\mathrm{time}$ records temporal information if available, $\mathrm{src}$ denotes source-related reliability information, and $\mathrm{doc}$ and $\mathrm{chunk}$ indicate the source document and chunk position. 
The source fields are retained as provenance information, allowing each hyperedge to be traced back to its original corpus location when needed.

Each extracted entity $v\in\widehat{\mathcal{V}}_{i,k}$ is represented as:
\begin{equation}
v
=
(\mathrm{name},\mathrm{type},\mathrm{role},\mathrm{alias},\mathrm{desc},\mathrm{conf}),
\label{eq:app_entity_record}
\end{equation}
where $\mathrm{name}$ is the entity name, $\mathrm{type}$ is the entity type, $\mathrm{role}$ describes its role in the corresponding knowledge segment, $\mathrm{alias}$ records possible aliases, $\mathrm{desc}$ provides a short textual description, and $\mathrm{conf}$ denotes the extraction confidence.

The local incidence relation is created when an entity participates in a hyperedge:
\begin{equation}
(e,v)\in\widehat{\mathcal{A}}_{i,k}
\quad
\text{if entity } v \text{ participates in hyperedge } e .
\label{eq:app_incidence_relation}
\end{equation}
Thus, a hyperedge can associate multiple entities simultaneously and preserve the higher-order structure of an extracted knowledge segment.

After local extraction, HyCE-RAG merges records from different chunks into global records. 
Let $\phi_{\mathcal{V}}(\cdot)$ and $\phi_{\mathcal{E}}(\cdot)$ denote the canonical mappings from local entity and hyperedge records to their merged global records:
\begin{equation}
\phi_{\mathcal{V}}:
\bigcup_{i=1}^{N}
\bigcup_{k=1}^{m_i}
\widehat{\mathcal{V}}_{i,k}
\rightarrow
\mathcal{V},
\qquad
\phi_{\mathcal{E}}:
\bigcup_{i=1}^{N}
\bigcup_{k=1}^{m_i}
\widehat{\mathcal{E}}_{i,k}
\rightarrow
\mathcal{E}.
\label{eq:app_canonical_mappings}
\end{equation}

The global entity set and hyperedge set are obtained as:
\begin{equation}
\mathcal{V}
=
\left\{
\phi_{\mathcal{V}}(\hat{v})
\mid
\hat{v}\in
\widehat{\mathcal{V}}_{i,k},
\ 1\leq i\leq N,\ 
1\leq k\leq m_i
\right\},
\label{eq:app_global_entities}
\end{equation}
\begin{equation}
\mathcal{E}
=
\left\{
\phi_{\mathcal{E}}(\hat{e})
\mid
\hat{e}\in
\widehat{\mathcal{E}}_{i,k},
\ 1\leq i\leq N,\ 
1\leq k\leq m_i
\right\}.
\label{eq:app_global_hyperedges}
\end{equation}

The global incidence relation is constructed by remapping local incidence links to their corresponding global records:
\begin{equation}
\mathcal{A}
=
\left\{
\left(
\phi_{\mathcal{E}}(\hat{e}),
\phi_{\mathcal{V}}(\hat{v})
\right)
\mid
(\hat{e},\hat{v})\in
\widehat{\mathcal{A}}_{i,k},
\ 1\leq i\leq N,\ 
1\leq k\leq m_i
\right\}.
\label{eq:app_global_incidence}
\end{equation}

The corpus-level hypergraph is therefore:
\begin{equation}
\mathcal{G}
=
(\mathcal{V},\mathcal{E},\mathcal{A}).
\label{eq:app_global_hypergraph}
\end{equation}

To construct the vector-based entity index, the textual representation of each merged entity $v\in\mathcal{V}$ is defined as:
\begin{equation}
r(v)
=
\operatorname{Concat}(\mathrm{name},\mathrm{alias},\mathrm{type},\mathrm{desc}),
\label{eq:app_entity_text_representation}
\end{equation}
and the corresponding embedding is computed by:
\begin{equation}
\mathbf{z}_v
=
f_{\mathrm{enc}}(r(v)).
\label{eq:app_offline_entity_embedding}
\end{equation}
These embeddings are stored in the entity index $\mathcal{I}_{\mathcal{V}}$ for vector-based entity retrieval during the online stage.

In implementation, each hyperedge is stored as a special hyperedge node and connected to its participating entity nodes through incidence links. 
This design allows the hypergraph to be stored and traversed using a graph database while preserving the higher-order structure of hyperedges.

\section{Convergence of Hypergraph Confidence Propagation}
\label{app:confidence_proof}

We show that the propagation process in Eq.~\eqref{eq:confidence_propagation} converges to a unique fixed point.

First, the transition matrix 
$\mathbf{S}=\mathbf{B}\mathbf{D}_{e}^{-1}\mathbf{B}^{\top}\mathbf{D}_{v}^{-1}$ 
is column-stochastic. 
Let $\mathbf{d}_{v}$ and $\mathbf{d}_{e}$ denote the vertex-degree and hyperedge-degree vectors, respectively. 
Since all isolated vertices and empty hyperedges are removed, all retained degrees are strictly positive. 
Using $\mathbf{1}^{\top}\mathbf{B}=\mathbf{d}_{e}^{\top}$ and 
$\mathbf{1}^{\top}\mathbf{B}^{\top}=\mathbf{d}_{v}^{\top}$, we have:
\begin{equation}
\begin{aligned}
\mathbf{1}^{\top}\mathbf{S}
&=
\mathbf{1}^{\top}
\mathbf{B}
\mathbf{D}_{e}^{-1}
\mathbf{B}^{\top}
\mathbf{D}_{v}^{-1} \\
&=
\mathbf{d}_{e}^{\top}
\mathbf{D}_{e}^{-1}
\mathbf{B}^{\top}
\mathbf{D}_{v}^{-1} \\
&=
\mathbf{1}^{\top}
\mathbf{B}^{\top}
\mathbf{D}_{v}^{-1}
=
\mathbf{d}_{v}^{\top}
\mathbf{D}_{v}^{-1}
=
\mathbf{1}^{\top}.
\end{aligned}
\label{eq:app_column_stochastic}
\end{equation}
Thus, each column of $\mathbf{S}$ sums to one. 
Because $\mathbf{S}$ is nonnegative and column-stochastic, its induced $\ell_1$ norm satisfies:
\begin{equation}
\|\mathbf{S}\|_1
=
\max_j\sum_i |\mathbf{S}_{ij}|
=
1.
\label{eq:app_s_norm}
\end{equation}
Therefore,
\begin{equation}
\rho(\mathbf{S})\leq \|\mathbf{S}\|_1=1,
\label{eq:app_spectral_radius}
\end{equation}
where $\rho(\cdot)$ denotes the spectral radius.

Next, recursively expanding Eq.~\eqref{eq:confidence_propagation} with $\mathbf{c}^{(0)}=\mathbf{r}$ gives:
\begin{equation}
\mathbf{c}^{(K)}
=
\lambda
\sum_{k=0}^{K-1}
(1-\lambda)^k
\mathbf{S}^{k}\mathbf{r}
+
(1-\lambda)^K
\mathbf{S}^{K}\mathbf{r}.
\label{eq:app_closed_form}
\end{equation}
The residual term vanishes as $K\rightarrow\infty$. 
Indeed, since $\|\mathbf{S}^{K}\|_1\leq \|\mathbf{S}\|_1^K=1$ and $\|\mathbf{r}\|_1=1$, we have:
\begin{equation}
\left\|
(1-\lambda)^K
\mathbf{S}^{K}\mathbf{r}
\right\|_1
\leq
(1-\lambda)^K.
\label{eq:app_residual_bound}
\end{equation}
For $\lambda\in(0,1]$, $0\leq 1-\lambda<1$, and hence:
\begin{equation}
\lim_{K\rightarrow\infty}
(1-\lambda)^K
=
0.
\label{eq:app_residual_limit}
\end{equation}
Thus, the residual term in Eq.~\eqref{eq:app_closed_form} disappears in the limit.

Moreover,
\begin{equation}
\rho((1-\lambda)\mathbf{S})
=
(1-\lambda)\rho(\mathbf{S})
\leq
1-\lambda
<
1.
\label{eq:app_scaled_radius}
\end{equation}
Therefore, the Neumann series converges:
\begin{equation}
\sum_{k=0}^{\infty}
\left((1-\lambda)\mathbf{S}\right)^k
=
\left(
\mathbf{I}
-
(1-\lambda)\mathbf{S}
\right)^{-1}.
\label{eq:app_neumann}
\end{equation}
Taking the limit of Eq.~\eqref{eq:app_closed_form}, we obtain:
\begin{equation}
\mathbf{c}^{\infty}
=
\lambda
\sum_{k=0}^{\infty}
(1-\lambda)^k
\mathbf{S}^{k}\mathbf{r}
=
\lambda
\left(
\mathbf{I}
-
(1-\lambda)\mathbf{S}
\right)^{-1}
\mathbf{r}.
\label{eq:app_limit}
\end{equation}

Finally, this limit is the unique fixed point. 
If $\mathbf{c}$ is a fixed point, then:
\begin{equation}
\mathbf{c}
=
\lambda\mathbf{r}
+
(1-\lambda)\mathbf{S}\mathbf{c}.
\label{eq:app_fixed_point}
\end{equation}
Rearranging gives:
\begin{equation}
\left(
\mathbf{I}
-
(1-\lambda)\mathbf{S}
\right)\mathbf{c}
=
\lambda\mathbf{r}.
\label{eq:app_fixed_linear}
\end{equation}
Since $\rho((1-\lambda)\mathbf{S})<1$, the matrix 
$\mathbf{I}-(1-\lambda)\mathbf{S}$ is invertible. 
Therefore, the fixed point is unique and equals:
\begin{equation}
\mathbf{c}
=
\lambda
\left(
\mathbf{I}
-
(1-\lambda)\mathbf{S}
\right)^{-1}
\mathbf{r}
=
\mathbf{c}^{\infty}.
\label{eq:app_unique_fixed_point}
\end{equation}

\section{Hyperparameter Settings}
\label{app:hyperparameters}

We use a fixed set of hyperparameters for HyCE-RAG across all test examples.
The core settings are summarized in Table~\ref{tab:hyce_hyperparameters}.
Unless otherwise specified, these hyperparameters are kept unchanged for all datasets and queries.

\begin{center}
\small
\begin{tabular}{l c}
\toprule
\textbf{Hyperparameter} & \textbf{Value} \\
\midrule
Number of entry entities & 5 \\
Expansion hops & 4 \\
Propagation steps & 10 \\
Selected evidence units & 15 \\
Final context evidence segments & 6 \\
\bottomrule
\end{tabular}
\captionof{table}{Core hyperparameter settings used for HyCE-RAG in our experiments.}
\label{tab:hyce_hyperparameters}
\end{center}

For each question, HyCE-RAG retrieves up to 5 entry entities and expands the hypergraph for 4 hops to collect query-relevant entities and hyperedges.
It then performs 10 confidence propagation steps over the query-aware hypergraph.
After evidence scoring and assembly, the top 15 evidence units are retained as the selected evidence set.
For answer generation, the final context is capped at 6 evidence segments to keep the evidence budget comparable across retrieval-based methods.

\bibliographystyle{cas-model2-names}
\bibliography{references}

@article{DBLP:journals/corr/abs-2002-08909,
  author       = {Kelvin Guu and
                  Kenton Lee and
                  Zora Tung and
                  Panupong Pasupat and
                  Ming{-}Wei Chang},
  title        = {{REALM:} Retrieval-Augmented Language Model Pre-Training},
  journal      = {CoRR},
  volume       = {abs/2002.08909},
  year         = {2020},
  url          = {https://arxiv.org/abs/2002.08909},
  eprinttype   = {arXiv},
  eprint       = {2002.08909},
  bibsource    = {dblp computer science bibliography, https://dblp.org}
}

@article{DBLP:journals/corr/abs-2312-10997,
  author       = {Yunfan Gao and
                  Yun Xiong and
                  Xinyu Gao and
                  Kangxiang Jia and
                  Jinliu Pan and
                  Yuxi Bi and
                  Yi Dai and
                  Jiawei Sun and
                  Qianyu Guo and
                  Meng Wang and
                  Haofen Wang},
  title        = {Retrieval-Augmented Generation for Large Language Models: {A} Survey},
  journal      = {CoRR},
  volume       = {abs/2312.10997},
  year         = {2023},
  url          = {https://doi.org/10.48550/arXiv.2312.10997},
  doi          = {10.48550/ARXIV.2312.10997},
  eprinttype   = {arXiv},
  eprint       = {2312.10997},
  bibsource    = {dblp computer science bibliography, https://dblp.org}
}

@article{zhao2026survey,
  title={A survey of large language models},
  author={Zhao, Wayne Xin and Zhou, Kun and Li, Junyi and Tang, Tianyi and Dong, Zican and Hou, Yupeng and Zhang, Beichen and Min, Yingqian and Zhang, Junjie and Liu, Peiyu and others},
  journal={Frontiers of Computer Science},
  volume={20},
  number={12},
  pages={2012627},
  year={2026},
  publisher={Springer}
}

@article{lewis2020retrieval,
  title={Retrieval-augmented generation for knowledge-intensive nlp tasks},
  author={Lewis, Patrick and Perez, Ethan and Piktus, Aleksandra and Petroni, Fabio and Karpukhin, Vladimir and Goyal, Naman and K{\"u}ttler, Heinrich and Lewis, Mike and Yih, Wen-tau and Rockt{\"a}schel, Tim and others},
  journal={Advances in neural information processing systems},
  volume={33},
  pages={9459--9474},
  year={2020}
}

@inproceedings{karpukhin2020dense,
  title={Dense passage retrieval for open-domain question answering},
  author={Karpukhin, Vladimir and Oguz, Barlas and Min, Sewon and Lewis, Patrick and Wu, Ledell and Edunov, Sergey and Chen, Danqi and Yih, Wen-tau},
  booktitle={Proceedings of the 2020 conference on empirical methods in natural language processing (EMNLP)},
  pages={6769--6781},
  year={2020}
}

@inproceedings{asai2024self,
  title={Self-rag: Learning to retrieve, generate, and critique through self-reflection},
  author={Asai, Akari and Wu, Zeqiu and Wang, Yizhong and Sil, Avi and Hajishirzi, Hannaneh},
  booktitle={International conference on learning representations},
  volume={2024},
  pages={9112--9141},
  year={2024}
}

@inproceedings{yang2018hotpotqa,
  title={HotpotQA: A dataset for diverse, explainable multi-hop question answering},
  author={Yang, Zhilin and Qi, Peng and Zhang, Saizheng and Bengio, Yoshua and Cohen, William and Salakhutdinov, Ruslan and Manning, Christopher D},
  booktitle={Proceedings of the 2018 conference on empirical methods in natural language processing},
  pages={2369--2380},
  year={2018}
}

@inproceedings{trivedi2023interleaving,
  title={Interleaving retrieval with chain-of-thought reasoning for knowledge-intensive multi-step questions},
  author={Trivedi, Harsh and Balasubramanian, Niranjan and Khot, Tushar and Sabharwal, Ashish},
  booktitle={Proceedings of the 61st annual meeting of the association for computational linguistics (volume 1: long papers)},
  pages={10014--10037},
  year={2023}
}

@inproceedings{jiang2023active,
  title={Active retrieval augmented generation},
  author={Jiang, Zhengbao and Xu, Frank F and Gao, Luyu and Sun, Zhiqing and Liu, Qian and Dwivedi-Yu, Jane and Yang, Yiming and Callan, Jamie and Neubig, Graham},
  booktitle={Proceedings of the 2023 conference on empirical methods in natural language processing},
  pages={7969--7992},
  year={2023}
}

@article{edge2024local,
  title={From local to global: A graph rag approach to query-focused summarization},
  author={Edge, Darren and Trinh, Ha and Cheng, Newman and Bradley, Joshua and Chao, Alex and Mody, Apurva and Truitt, Steven and Metropolitansky, Dasha and Ness, Robert Osazuwa and Larson, Jonathan},
  journal={arXiv preprint arXiv:2404.16130},
  year={2024}
}

@article{pan2024unifying,
  title={Unifying large language models and knowledge graphs: A roadmap},
  author={Pan, Shirui and Luo, Linhao and Wang, Yufei and Chen, Chen and Wang, Jiapu and Wu, Xindong},
  journal={IEEE Transactions on Knowledge and Data Engineering},
  volume={36},
  number={7},
  pages={3580--3599},
  year={2024},
  publisher={IEEE}
}

@article{luo2026hypergraphrag,
  title={Hypergraphrag: Retrieval-augmented generation via hypergraph-structured knowledge representation},
  author={Luo, Haoran and Chen, Guanting and Zheng, Yandan and Wu, Xiaobao and Guo, Yikai and Lin, Qika and Feng, Yu and Kuang, Zemin and Song, Meina and Zhu, Yifan and others},
  journal={Advances in Neural Information Processing Systems},
  volume={38},
  pages={152206--152234},
  year={2026}
}

@article{arslan2024survey,
  title={A Survey on RAG with LLMs},
  author={Arslan, Muhammad and Ghanem, Hussam and Munawar, Saba and Cruz, Christophe},
  journal={Procedia computer science},
  volume={246},
  pages={3781--3790},
  year={2024},
  publisher={Elsevier}
}

@article{gupta2024comprehensive,
  title={A comprehensive survey of retrieval-augmented generation (rag): Evolution, current landscape and future directions},
  author={Gupta, Shailja and Ranjan, Rajesh and Singh, Surya Narayan},
  journal={arXiv preprint arXiv:2410.12837},
  year={2024}
}

@article{xiang2025use,
  title={When to use graphs in rag: A comprehensive analysis for graph retrieval-augmented generation},
  author={Xiang, Zhishang and Wu, Chuanjie and Zhang, Qinggang and Chen, Shengyuan and Hong, Zijin and Huang, Xiao and Su, Jinsong},
  journal={arXiv preprint arXiv:2506.05690},
  year={2025}
}

@inproceedings{zeeshan2025rag,
  title={Rag powered llms for qa: Evolution, challenges, applications, and future directions},
  author={Zeeshan, Hafiz Muhammad Ali and Faizan, Muhammad and Zia, Usman and Gohar, Abdullah},
  booktitle={2025 International Conference on Communication Technologies (ComTech)},
  pages={1--6},
  year={2025},
  organization={IEEE}
}

@inproceedings{jiang2025retrieve,
  title={Retrieve, summarize, plan: Advancing multi-hop question answering with an iterative approach},
  author={Jiang, Zhouyu and Sun, Mengshu and Liang, Lei and Zhang, Zhiqiang},
  booktitle={Companion Proceedings of the ACM on Web Conference 2025},
  pages={1677--1686},
  year={2025}
}

@inproceedings{zhu2025knowledge,
  title={Knowledge graph-guided retrieval augmented generation},
  author={Zhu, Xiangrong and Xie, Yuexiang and Liu, Yi and Li, Yaliang and Hu, Wei},
  booktitle={Proceedings of the 2025 Conference of the Nations of the Americas Chapter of the Association for Computational Linguistics: Human Language Technologies (Volume 1: Long Papers)},
  pages={8912--8924},
  year={2025}
}

@article{guo2024lightrag,
  title={Lightrag: Simple and fast retrieval-augmented generation},
  author={Guo, Zirui and Xia, Lianghao and Yu, Yanhua and Ao, Tian and Huang, Chao},
  journal={arXiv preprint arXiv:2410.05779},
  volume={2},
  number={3},
  year={2024}
}

@article{luo2026gfm,
  title={GFM-RAG: graph foundation model for retrieval augmented generation},
  author={Luo, Linhao and Zhao, Zicheng and Haffari, Reza and Phung, Dinh and Gong, Chen and Pan, Shirui},
  journal={Advances in Neural Information Processing Systems},
  volume={38},
  pages={36371--36405},
  year={2026}
}

@inproceedings{xu2024multi,
  title={Multi-perspective improvement of knowledge graph completion with large language models},
  author={Xu, Derong and Zhang, Ziheng and Lin, Zhenxi and Wu, Xian and Zhu, Zhihong and Xu, Tong and Zhao, Xiangyu and Zheng, Yefeng and Chen, Enhong},
  booktitle={Proceedings of the 2024 joint international conference on computational linguistics, language resources and evaluation (LREC-COLING 2024)},
  pages={11956--11968},
  year={2024}
}

@inproceedings{ho2020constructing,
  title={Constructing a multi-hop qa dataset for comprehensive evaluation of reasoning steps},
  author={Ho, Xanh and Nguyen, Anh-Khoa Duong and Sugawara, Saku and Aizawa, Akiko},
  booktitle={Proceedings of the 28th International Conference on Computational Linguistics},
  pages={6609--6625},
  year={2020}
}

@article{trivedi2022musique,
  title={♫ MuSiQue: Multihop Questions via Single-hop Question Composition},
  author={Trivedi, Harsh and Balasubramanian, Niranjan and Khot, Tushar and Sabharwal, Ashish},
  journal={Transactions of the Association for Computational Linguistics},
  volume={10},
  pages={539--554},
  year={2022},
  publisher={MIT Press One Broadway, 12th Floor, Cambridge, Massachusetts 02142, USA~…}
}

@article{mavi2024multi,
  title={Multi-hop question answering},
  author={Mavi, Vaibhav and Jangra, Anubhav and Jatowt, Adam},
  journal={Foundations and Trends{\textregistered} in Information Retrieval},
  volume={17},
  number={4},
  pages={457--586},
  year={2024},
  publisher={Emerald Publishing Limited}
}

@article{chen2019multi,
  title={Multi-hop question answering via reasoning chains},
  author={Chen, Jifan and Lin, Shih-ting and Durrett, Greg},
  journal={arXiv preprint arXiv:1910.02610},
  year={2019}
}

@article{xiong2020answering,
  title={Answering complex open-domain questions with multi-hop dense retrieval},
  author={Xiong, Wenhan and Li, Xiang Lorraine and Iyer, Srini and Du, Jingfei and Lewis, Patrick and Wang, William Yang and Mehdad, Yashar and Yih, Wen-tau and Riedel, Sebastian and Kiela, Douwe and others},
  journal={arXiv preprint arXiv:2009.12756},
  year={2020}
}

@article{asai2019learning,
  title={Learning to retrieve reasoning paths over wikipedia graph for question answering},
  author={Asai, Akari and Hashimoto, Kazuma and Hajishirzi, Hannaneh and Socher, Richard and Xiong, Caiming},
  journal={arXiv preprint arXiv:1911.10470},
  year={2019}
}

@article{yao2022react,
  title={React: Synergizing reasoning and acting in language models},
  author={Yao, Shunyu and Zhao, Jeffrey and Yu, Dian and Du, Nan and Shafran, Izhak and Narasimhan, Karthik and Cao, Yuan},
  journal={arXiv preprint arXiv:2210.03629},
  year={2022}
}

@article{xiao2025graphrag,
  title={Graphrag-bench: Challenging domain-specific reasoning for evaluating graph retrieval-augmented generation},
  author={Xiao, Yilin and Dong, Junnan and Zhou, Chuang and Dong, Su and Zhang, Qian-wen and Yin, Di and Sun, Xing and Huang, Xiao},
  journal={arXiv preprint arXiv:2506.02404},
  year={2025}
}

@inproceedings{roberts2020much,
  title={How much knowledge can you pack into the parameters of a language model?},
  author={Roberts, Adam and Raffel, Colin and Shazeer, Noam},
  booktitle={Proceedings of the 2020 conference on empirical methods in natural language processing (EMNLP)},
  pages={5418--5426},
  year={2020}
}

@article{zhuang2025linearrag,
  title={LinearRAG: Linear Graph Retrieval Augmented Generation on Large-scale Corpora},
  author={Zhuang, Luyao and Chen, Shengyuan and Xiao, Yilin and Zhou, Huachi and Zhang, Yujing and Chen, Hao and Zhang, Qinggang and Huang, Xiao},
  journal={arXiv preprint arXiv:2510.10114},
  year={2025}
}

@misc{openai2025gptoss120bgptoss20bmodel,
      title={gpt-oss-120b \& gpt-oss-20b Model Card}, 
      author={OpenAI},
      year={2025},
      eprint={2508.10925},
      archivePrefix={arXiv},
      primaryClass={cs.CL},
      url={https://arxiv.org/abs/2508.10925}, 
}

@inproceedings{kwon2023efficient,
  title={Efficient Memory Management for Large Language Model Serving with PagedAttention},
  author={Woosuk Kwon and Zhuohan Li and Siyuan Zhuang and Ying Sheng and Lianmin Zheng and Cody Hao Yu and Joseph E. Gonzalez and Hao Zhang and Ion Stoica},
  booktitle={Proceedings of the ACM SIGOPS 29th Symposium on Operating Systems Principles},
  year={2023}
}

@inproceedings{zhao2017algorithm,
  title={An algorithm for the orientation of complete bipartite graphs},
  author={Zhao, Lingqi and Wang, Mujiangshan and Zhang, Xuefei and Lin, Yuqing and Wang, Shiying},
  booktitle={2017 International Conference on Applied Mathematics, Modelling and Statistics Application (AMMSA 2017)},
  pages={361--364},
  year={2017},
  organization={Atlantis Press}
}

@article{wang2016diagnosability,
  title={Diagnosability of Cayley graph networks generated by transposition trees under the comparison diagnosis model},
  author={Wang, Mujiangshan and Wang, Shiying},
  journal={Annals of Applied Mathematics},
  volume={32},
  number={2},
  pages={166--173},
  year={2016}
}

@article{wang2018edge,
  title={The Edge Connectivity of Expanded k-Ary n-Cubes},
  author={Wang, Shiying and Wang, Mujiangshan},
  journal={Discrete Dynamics in Nature and Society},
  volume={2018},
  number={1},
  pages={7867342},
  year={2018},
  publisher={Wiley Online Library}
}

@article{mu2010ordered,
  title={Ordered and Hamilton Digraphs},
  author={Mu-Jiang-shan, WANG and Jun, YUAN and Shang-wei, LIN and others},
  journal={Chinese Quarterly Journal of Mathematics},
  volume={25},
  number={3},
  pages={317--326},
  year={2010}
}

@article{wei2025fstgat,
  title={FSTGAT: Financial Spatio-Temporal Graph Attention Network for Non-Stationary Financial Systems and Its Application in Stock Price Prediction},
  author={Wei, Ze-Lin and An, Hong-Yu and Yao, Yao and Su, Wei-Cong and Li, Guo and Saifullah and Sun, Bi-Feng and Wang, Mu-Jiang-Shan},
  journal={Symmetry},
  volume={17},
  number={8},
  pages={1344},
  year={2025},
  publisher={MDPI}
}

@inproceedings{wang2011embedding,
  title={Embedding paths into the 4-ary n-cube with faulty nodes},
  author={Wang, Shiying and Wangmu, Jiangshan and Qi, Zhifang and Ren, Yunxia},
  booktitle={2011 International Conference on Consumer Electronics, Communications and Networks (CECNet)},
  pages={4949--4951},
  year={2011},
  organization={IEEE}
}

@article{jian2026pi,
  title={PI-VLA: Adaptive Symmetry-Aware Decision-Making for Long-Horizon Vision--Language--Action Manipulation},
  author={Jian, Yina and Tian, Di and Chen, Xuan-Jing and Wei, Zhen-Yuan and Liang, Chen-Wei and Wang, Mu-Jiang-Shan},
  journal={Symmetry},
  volume={18},
  number={3},
  pages={394},
  year={2026},
  publisher={MDPI}
}



\end{document}